\pdfoutput=1

\documentclass[11pt]{article}

\usepackage[final]{acl}
\usepackage{booktabs}

\usepackage{times}
\usepackage{latexsym}
\usepackage{array}       %
\usepackage{pdflscape}  
\usepackage{geometry}
\usepackage{longtable}
\usepackage{enumitem}
\usepackage{float}
\usepackage{caption}
\usepackage{float}    
\usepackage{placeins} 
\usepackage[T1]{fontenc}
\usepackage{multirow}

\usepackage[utf8]{inputenc}

\usepackage{microtype}

\usepackage{inconsolata}

\usepackage{graphicx}

\usepackage{xspace}

\newcommand{\methodname}{\textsc{EnDive}\xspace}

\title{EnDive: A Cross-Dialect Benchmark for Fairness and Performance in Large Language Models}

\author{%
Abhay Gupta\thanks{Lead Author} \quad Jacob Cheung \quad Philip Meng \quad Shayan Sayyed
\\
\textbf{Austen Liao}\thanks{Senior Author} \quad \textbf{Kevin Zhu}\footnotemark[2] \quad \textbf{Sean O'Brien}\footnotemark[2]
\\
Algoverse AI Research\\
\texttt{abhaygupta1266@gmail.com, kevin@algoverse.us}
}

\begin{document}
\maketitle

\begin{abstract}
The diversity of human language, shaped by social, cultural, and regional influences, presents significant challenges for natural language processing (NLP) systems. Existing benchmarks often overlook intra-language variations, leaving speakers of non-standard dialects underserved. To address this gap, we introduce \textbf{\methodname} (\textbf{En}glish \textbf{Div}ersity), a benchmark that evaluates five widely-used large language models (LLMs) across tasks in language understanding, algorithmic reasoning, mathematics, and logic. Our framework translates Standard American English datasets into five underrepresented dialects using few-shot prompting with verified examples from native speakers, and compare these translations against rule-based methods via fluency assessments, preference tests, and semantic similarity metrics. Human evaluations confirm high translation quality, with average scores of at least 6.02/7 for faithfulness, fluency, and formality. By filtering out near-identical translations, we create a challenging dataset that reveals significant performance disparities—\textbf{models consistently underperform on dialectal inputs compared to Standard American English.} \textbf{\methodname} thus advances dialect-aware NLP by uncovering model biases and promoting more equitable language technologies.
\end{abstract}

\section{Introduction}
Language diversity, shaped by social and cultural factors, presents significant challenges for NLP systems. While English serves as a global lingua franca, its dialects exhibit substantial variation that often goes unaddressed in language technologies \cite{chambers1998dialectology}. This oversight perpetuates discrimination against dialect speakers in critical domains like education and employment \cite{purnell1999dialect, hofmann2024ai_racist_decisions}, exacerbated by LLMs'
predominant focus on Standard American English (SAE) \cite{blodgett-etal-2016-demographic}.

Recent studies reveal systemic biases in LLM processing of non-standard dialects \cite{fleisig2024linguisticbiaschatgptlanguage, resende2024comprehensiveviewbiasestoxicity}—from toxic speech misclassification of African American Vernacular English tweets \cite{sap-etal-2019-risk} to parsing errors in Chicano and Jamaican English \cite{fought2003chicano, patrick1999urban}. Similar issues plague Indian and Singaporean English due to morphological divergences \cite{kachru1983indianization, gupta1994step}, highlighting an urgent need for inclusive NLP systems \cite{ziems-etal-2022-value}.

Existing benchmarks like GLUE \cite{wang2019gluemultitaskbenchmarkanalysis} and SuperGLUE \cite{wang2020supergluestickierbenchmarkgeneralpurpose} fail to capture dialect variation, while specialized datasets (SVAMP, MBPP, FOLIO) \cite{patel-etal-2021-nlp, austin2021programsynthesislargelanguage, han2024folionaturallanguagereasoning} remain SAE-centric. While frameworks like Multi-VALUE \cite{ziems2023multivalueframeworkcrossdialectalenglish, ziems-etal-2022-value} address dialect representation through rule-based lexical substitutions, their synthetic approach fails to capture authentic syntactic patterns. This limitation is particularly acute in reasoning tasks, where surface-level translations preserve logical meaning but lose dialect-specific pragmatic markers essential for fair evaluation.

To address these gaps, we introduce \textbf{\methodname} (\textbf{En}glish \textbf{Dive}rsity), a benchmark that evaluates five LLMs across 12 natural language understanding (NLU) tasks translated into five underrepresented dialects selected for their linguistic distinctiveness and sociocultural significance:

\vspace{0.1em} 

\begin{itemize}[nosep,leftmargin=*]
    \item \textbf{African American Vernacular English (AAVE)}: 33M speakers with distinct syntax/phonology \cite{lippi1997english}
    \item \textbf{Indian English (IndE)}: 250M speakers blending local/colonial influences \cite{kachru1983indianization}
    \item \textbf{Jamaican English (JamE)}: Diaspora language with mesolectal variation \cite{patrick1999urban}
    \item \textbf{Chicano English (ChcE)}: Spanish-influenced variety in US Hispanic communities \cite{fought2003chicano}
    \item \textbf{Colloquial Singaporean English (CollSgE)}: Multicultural creole with Asian substrates \cite{platt1980english}
\end{itemize}

\vspace{0.1em} 

Our methodology combines linguistic authenticity with strategic filtering to create robust dialect evaluations. Using verified text samples in the target dialects from eWAVE \cite{ewave} for few-shot prompting, we translate SAE datasets into target dialects while preserving sociolinguistic nuance. To eliminate superficial transformations, we apply BLEU-based filtering \cite{papineni-etal-2002-bleu}, removing translations with scores $\geq$0.7 against their SAE sources—retaining only substantive linguistic variations that challenge LLMs' dialect understanding. We compare our translations against Multi-VALUE's rule-based translations \cite{ziems2023multivalueframeworkcrossdialectalenglish} through fluency assessments, semantic similarity metrics, and LLM preference tests. Additionally, we have native speakers assess our translations to ensure linguistic authenticity and original content meaning are preserved across all five dialects.

\vspace{0.1em} 

\noindent\textbf{Our Contributions:}
\begin{enumerate}[nosep,leftmargin=*,label=\textbf{(\arabic*)}]
    \item \textbf{Public Benchmark}: Curated challenging dialectal variants across 12 reasoning and natural language understanding tasks validated for translation fidelity several metrics and human validation.
    
    \item \textbf{Cross-LLM Evaluation}: Comprehensive testing of 5 LLMs (GPT-4o, GPT-4o mini, Claude-3.5-Sonnet, Deepseek-v3, LLaMa-3-8b) revealing consistent performance disparities between SAE and dialectal inputs using chain-of-thought (CoT) and zero-shot prompting.
\end{enumerate}

\section{Related Work}

\textbf{Dialectal Diversity.} Addressing dialectal diversity in NLP remains a significant challenge due to inherent linguistic variations shaped by social and cultural contexts. Early research identified systemic biases in language models against non-standard dialects such as AAVE, highlighting issues like the misclassification of AAVE tweets as toxic and difficulties in syntactic parsing \cite{sap-etal-2019-risk, jorgensen-etal-2015-challenges}. Recent studies extend these findings to modern LLMs, revealing persistent dialect prejudice in evaluations related to employability, criminality, and medical diagnoses \cite{hofmann2024dialectprejudicepredictsai, fleisig2024linguisticbiaschatgptlanguage, blodgett2017racialdisparitynaturallanguage}.

\textbf{Benchmarking Approaches.} Benchmarking dialect robustness has primarily followed two approaches. The first employs rule-based lexical substitutions in frameworks like VALUE and Multi-VALUE \cite{ziems-etal-2022-value, ziems2023multivalueframeworkcrossdialectalenglish}. While scalable, these methods often fail to capture nuanced, context-dependent linguistic features essential for authentic dialect representation, such as AAVE’s habitual “be” \cite{green2002african, lippi1997english} or Chicano English’s Spanish-influenced prosody \cite{fought2003chicano, santa1993chicano}. The second approach relies on human-annotated translations for authenticity, as seen in datasets like ReDial and AraDiCE \cite{lin2025languagegapsevaluatingdialect, mousi2024aradicebenchmarksdialectalcultural}, but these typically focus on single dialects, limiting their applicability for comprehensive dialect fairness evaluations across multiple linguistic variations.

\textbf{Hybrid Human-Machine Methodologies.} Emerging hybrid approaches combine automated translation techniques with human validation to mitigate the limitations of purely rule-based or human-annotated methods. For example, AraDiCE \cite{mousi2024aradicebenchmarksdialectalcultural} integrates automated translations with native speaker post-edits for Arabic dialects, while ReDial \cite{lin2025languagegapsevaluatingdialect} leverages human validation to ensure cultural and linguistic fidelity. Similarly, AAVENUE \cite{gupta2024aavenuedetectingllmbiases} offers human-validated evaluations for AAVE in NLU tasks but remains restricted to a single dialect.

\textbf{Sociolinguistic Impact and Real-World Discrimination.} Beyond technical benchmarks, sociolinguistic studies have linked LLM biases to real-world discrimination—such as housing denials for AAVE speakers \cite{hofmann2024dialectprejudicepredictsai, purnell1999dialect} and biased criminal justice assessments \cite{fleisig2024linguisticbiaschatgptlanguage}. Multilingual initiatives like LLM for Everyone \cite{cahyawijaya2024llmeveryonerepresentingunderrepresented} advocate for continuous tuning of models to improve performance on underrepresented languages, an approach that aligns with our use of human-guided few-shot prompting informed by authentic linguistic examples \cite{ewave, platt1980english}.

\textbf{Remaining Gaps and Our Contribution.} Although prior work has deepened our understanding of dialect biases in NLP, significant gaps remain in developing comprehensive, multi-dialect benchmarks that integrate authentic linguistic features. \textbf{\methodname} addresses these gaps by providing a robust benchmark that combines both automated and human-validated translation methods, thereby fostering more equitable language technology development.

\section{Dataset}

\subsection{Dataset Overview}
\textbf{\methodname} is a benchmark designed to evaluate the reasoning capabilities of LLMs across five underrepresented dialects. The benchmark is curated from 12 established datasets, spanning four core reasoning categories: \textcolor[rgb]{0.2,0.4,0.8}{\textbf{Language Understanding}}, \textcolor[rgb]{0.6,0.2,0.6}{\textbf{Algorithmic Understanding}}, \textcolor[rgb]{0.2,0.6,0.4}{\textbf{Math}}, and \textcolor[rgb]{0.8,0.4,0.2}{\textbf{Logic}}. Tasks were translated from SAE into the target dialects using few-shot prompting informed by eWAVE examples. For comparison, we generate parallel translations using Multi-VALUE's rule-based framework.

\subsection{Data Sourcing}
The dataset comprises tasks selected from diverse and established benchmarks. Below, we describe each dataset, its focus, and the sampled instances.

\paragraph{\textcolor[rgb]{0.2,0.4,0.8}{Language Understanding}} 
\textbf{BoolQ} \cite{wang2020supergluestickierbenchmarkgeneralpurpose} is a yes/no question-answering task derived from Wikipedia passages, testing the model’s ability to determine factual correctness. We sampled 1,000 instances.  
\textbf{MultiRC} \cite{wang2020supergluestickierbenchmarkgeneralpurpose} requires multi-sentence reasoning with each question having multiple correct answers. We included 1,000 examples.  
\textbf{WSC} \cite{wang2020supergluestickierbenchmarkgeneralpurpose} assesses coreference resolution, requiring commonsense knowledge to match pronouns with their correct referents. We included 659 examples.  
\textbf{SST-2} \cite{wang2019gluemultitaskbenchmarkanalysis} evaluates binary sentiment classification on movie reviews, labeling each as positive or negative. A total of 1,000 instances were included.  
\textbf{COPA} \cite{wang2020supergluestickierbenchmarkgeneralpurpose} is a causal reasoning task where models identify the correct cause or effect from two choices. We included 500 examples.

\paragraph{\textcolor[rgb]{0.6,0.2,0.6}{Algorithmic Understanding}}
\textbf{HumanEval} \cite{chen2021evaluatinglargelanguagemodels} is a benchmark of human-crafted Python coding problems, each paired with test cases to evaluate correctness. We sampled 164 examples.  
\textbf{MBPP} \cite{austin2021programsynthesislargelanguage} contains Python coding tasks designed for program synthesis and correctness evaluation. A total of 374 examples were included.

\paragraph{\textcolor[rgb]{0.2,0.6,0.4}{Math}} 
\textbf{GSM8K} \cite{cobbe2021trainingverifierssolvemath} presents grade-school math word problems requiring numeric reasoning and problem-solving. We included 1,000 examples.  
\textbf{SVAMP} \cite{patel-etal-2021-nlp} features systematically modified arithmetic problems that test robustness in mathematical reasoning. We sampled 700 examples.

\paragraph{\textcolor[rgb]{0.8,0.4,0.2}{Logic}} 
\textbf{LogicBench} \cite{parmar2024logicbenchsystematicevaluationlogical} comprises logical reasoning tasks in both Yes/No and multiple-choice formats, designed to evaluate deductive reasoning capabilities. A total of 980 examples were included, with 500 instances from Yes/No tasks and 480 from multiple-choice tasks.  
\textbf{FOLIO} \cite{han2024folionaturallanguagereasoning} features first-order logic challenges presented in natural language, requiring models to identify valid conclusions or contradictions. We sampled 1,000 examples for this task.

\subsection{Few-Shot Prompting for Dialect Translation}
To translate tasks from SAE into each of the five underrepresented dialects, we employed a few-shot prompting strategy \cite{brown2020languagemodelsfewshotlearners} informed by examples from eWAVE \cite{ewave}, a linguistically validated resource that documents and analyzes structural variations across global English dialects. We utilized three utlized exemplar translations from eWAVE per dialect. Utilizing GPT-4o \cite{openai2024gpt4technicalreport}, the language model was then prompted to rewrite the input text in the desired dialect based on these exemplars. This approach ensures that translations maintain linguistic authenticity and accurately reflect the sociocultural nuances inherent to each dialect. Detailed examples of these prompts can be found in \textbf{Section~\ref{sec:appendix}} in the appendix.

\subsection{Baseline Translations with Multi-VALUE}
To establish a baseline for comparison, we generated translations using Multi-VALUE \cite{ziems2023multivalueframeworkcrossdialectalenglish}, a rule-based framework designed to produce synthetic dialectal transformations. Multi-VALUE applies predefined linguistic rules to transform SAE into target dialects, providing a systematic approach for generating dialectal variations.

The percentage of successful translations for each dataset and dialect is detailed in \textbf{Appendix~\ref{sec:translationmultiva}}, which highlights the variability in Multi-VALUE's performance. This underscores the necessity for more robust and context-aware translation methods, such as our few-shot prompting approach with GPT-4o.

\subsection{BLEU Score Filtering for Challenging Translations}

To create a more challenging benchmark, we applied BLEU score \cite{papineni-etal-2002-bleu} filtering to exclude translations with BLEU scores above 0.7, as these were overly similar to the original SAE text. This retained translations with greater linguistic diversity and structural differences, enhancing the benchmark’s focus on real-world dialectal variations. Detailed statistics on filtered translations are presented in \textbf{Appendix~\ref{sec:bleu_filtering_statistics}}. 

\section{Analysis}
\label{sec:analysis}
\subsection{ROUGE Diversity Score Evaluation}
\label{sec:rouge_score_evaluation}

\textbf{ROUGE Diversity} \cite{lin-2004-rouge}, calculated as the average of ROUGE-1, ROUGE-2, and ROUGE-L, measures lexical variation while preserving meaning. As detailed in Appendix~\ref{sec:metrics_table}, \textbf{\methodname} generally outperformed \textbf{Multi-VALUE}. For example, in SVAMP IndE, it scored 0.8418 vs. 0.7632, and in CollSgE MBPP, 0.7088 vs. 0.6181. However, in AAVE BoolQ, \textbf{Multi-VALUE} scored higher, suggesting occasional advantages in lexical overlap.

\subsection{Lexical Diversity Evaluation} 
\label{sec:lexical_diversity_evaluation}

Lexical diversity, which measures how varied the vocabulary is in a text, captures how well translations preserve the nuances of each dialect. As shown in Appendix~\ref{sec:metrics_table}, \textbf{\methodname} generally outperformed \textbf{Multi-VALUE}, achieving higher scores in most dialects and datasets. For example, in AAVE COPA, it scored 0.9864 vs. 0.9851, and in IndE GSM8K, 0.7237 vs. 0.7230. However, in JamE MBPP, Multi-VALUE scored higher (0.7370 vs. 0.6289), indicating occasional advantages. These results demonstrate \textbf{\methodname}'s effectiveness in maintaining lexical diversity across dialects.

\begin{table*}[t]
\centering
\footnotesize
\setlength{\tabcolsep}{3pt}
\renewcommand{\arraystretch}{1.1}
\resizebox{\textwidth}{!}{%
\begin{tabular}{lcccc|cccc|cccc|cccc|cccc}
\toprule
\multirow{2}{*}{\textbf{Dataset}} & 
\multicolumn{4}{c|}{\textbf{AAVE}} & 
\multicolumn{4}{c|}{\textbf{ChcE}} & 
\multicolumn{4}{c|}{\textbf{CollSgE}} & 
\multicolumn{4}{c|}{\textbf{IndE}} & 
\multicolumn{4}{c}{\textbf{JamE}} \\
\cmidrule(lr){2-5} \cmidrule(lr){6-9} \cmidrule(lr){10-13} \cmidrule(lr){14-17} \cmidrule(lr){18-21}
& \textit{ZS} & \textit{CoT} & \textit{SAE ZS} & \textit{SAE CoT} 
& \textit{ZS} & \textit{CoT} & \textit{SAE ZS} & \textit{SAE CoT} 
& \textit{ZS} & \textit{CoT} & \textit{SAE ZS} & \textit{SAE CoT} 
& \textit{ZS} & \textit{CoT} & \textit{SAE ZS} & \textit{SAE CoT} 
& \textit{ZS} & \textit{CoT} & \textit{SAE ZS} & \textit{SAE CoT} \\
\midrule
BoolQ          & 90.29 & 90.05 & 91.47 & \textbf{91.92} & 89.74 & 89.89 & 91.25 & \textbf{91.61} & 89.89 & 89.79 & 91.53 & \textbf{91.78} & 90.75 & 90.50 & 91.62 & \textbf{91.95} & 89.65 & 89.45 & 91.58 & \textbf{91.83} \\
COPA           & 97.16 & 96.93 & 96.77 & \textbf{97.42} & 96.88 & 96.47 & 97.20 & \textbf{97.45} & 97.33 & 97.33 & 97.10 & \textbf{97.40} & \textbf{98.10} & 98.10 & 97.36 & 97.81 & 94.59 & 94.99 & 97.01 & \textbf{97.37} \\
FOLIO          & 62.27 & 63.57 & 73.61 & \textbf{74.15} & 63.68 & 62.88 & 73.80 & \textbf{74.20} & 65.62 & 65.21 & 73.91 & \textbf{74.43} & 68.12 & 68.12 & 73.74 & \textbf{74.57} & 65.56 & 65.16 & 73.83 & \textbf{74.49} \\
GSM8K          & 60.86 & 84.05 & 89.54 & \textbf{90.27} & 59.54 & 77.17 & 89.25 & \textbf{90.10} & 51.28 & 78.40 & 89.38 & \textbf{90.19} & 60.36 & 87.13 & 89.41 & \textbf{90.32} & 60.07 & 80.86 & 89.29 & \textbf{90.22} \\
HumanEVAL      & 92.31 & 92.31 & 94.10 & \textbf{93.85} & \textbf{97.09} & 96.12 & 94.32 & 93.78 & 92.11 & 96.05 & 94.20 & \textbf{93.91} & 96.00 & 96.00 & 94.05 & 93.87 & 91.46 & 91.46 & 94.14 & \textbf{93.96} \\
SVAMP          & 92.67 & 90.99 & 94.11 & \textbf{94.51} & 92.77 & 91.96 & 94.05 & \textbf{94.40} & 92.46 & 90.63 & 94.22 & \textbf{94.54} & 92.77 & 91.58 & 94.09 & \textbf{94.48} & 92.99 & 90.11 & 94.18 & \textbf{94.47} \\
LogicBenchMCQ  & 78.41 & 73.96 & 82.52 & \textbf{83.65} & 79.58 & 73.85 & 82.48 & \textbf{83.70} & 80.38 & 73.54 & 82.60 & \textbf{83.57} & 79.83 & 74.48 & 82.50 & \textbf{83.74} & 78.87 & 72.92 & 82.66 & \textbf{83.71} \\
LogicBenchYN   & 77.45 & 76.12 & 75.63 & \textbf{76.97} & 76.69 & 75.56 & 75.51 & \textbf{76.83} & 77.44 & 75.40 & 75.74 & \textbf{76.92} & 78.06 & 76.02 & 75.55 & \textbf{76.91} & 77.21 & 75.69 & 75.66 & \textbf{76.78} \\
\textbf{MBPP}  & 85.29 & \textbf{86.49} & 85.92 & 74.31 & \textbf{86.73} & 85.80 & 85.84 & 74.17 & \textbf{86.98} & 85.50 & 85.95 & 74.35 & 84.00 & 83.00 & \textbf{85.79} & 74.42 & \textbf{86.92} & \textbf{86.92} & 85.86 & 74.38 \\
MultiRC        & 86.92 & 86.41 & 89.07 & \textbf{89.76} & 86.50 & 87.10 & 89.13 & \textbf{89.67} & 87.26 & 86.75 & 89.10 & \textbf{89.79} & 86.44 & 85.11 & 89.15 & \textbf{89.71} & 87.20 & 87.10 & 89.20 & \textbf{89.73} \\
WSC            & 54.83 & 51.55 & 81.69 & \textbf{88.42} & 54.95 & 50.53 & 81.55 & \textbf{88.29} & 54.71 & 51.54 & 81.71 & \textbf{88.39} & 62.57 & 53.82 & 81.49 & \textbf{88.41} & 54.23 & 53.19 & 81.61 & \textbf{88.47} \\
SST-2          & 91.91 & 92.25 & 89.97 & \textbf{93.12} & 91.62 & 91.30 & 89.80 & \textbf{93.04} & 90.06 & 89.64 & 89.94 & \textbf{93.19} & 91.08 & 90.95 & 89.86 & \textbf{93.08} & 89.55 & 89.01 & 89.82 & \textbf{93.10} \\
\bottomrule
\end{tabular}
}
\caption{DeepSeek-v3 Accuracy (\%). \textbf{Bold} indicates superior performance within dialect pairs.}
\label{tab:deepseek_dialect_sae_specific}
\end{table*}

\subsection{Fluency Evaluation}
\label{sec:fluency_evaluation}

Building upon our assessments of semantic alignment and lexical diversity, fluency evaluation ensures that translations are not only accurate but also natural and grammatically correct within the target dialect. Automatic fluency metrics are typically designed for SAE, making them less effective for dialectal translations. To address this, we use \textbf{GPT-4o} \cite{openai2024gpt4technicalreport} for fluency scoring, following prior work \cite{kocmi2023gembamqmdetectingtranslationquality} that leveraged LLMs for translation quality assessment. Our approach employs a detailed prompt in Appendix~\ref{sec:promptingfluncy} and \textbf{CoT} reasoning to ensure a structured evaluation. As shown in Appendix~\ref{sec:metrics_table}, \textbf{\methodname} achieves consistently high fluency scores across dialects on a 1-7 scale, with higher scores indicating greater fluency. Notably, AAVE COPA and AAVE MultiRC scored 6.83, reflecting strong alignment with dialectal norms. Similarly, JamE HumanEVAL achieved 6.45, indicating natural fluency in Jamaican English.

\subsection{Preference Tests}
\label{sec:preference_evaluation}

Pairwise preference tests were conducted to compare \textbf{\methodname} and \textbf{Multi-VALUE} translations using \textbf{GPT-4o} with \textbf{CoT}. The prompt, detailed in Appendix~\ref{sec:prefscoreprom}, evaluated translations based on fluency, accuracy, readability, and cultural appropriateness. As shown in Appendix~\ref{sec:metrics_table}, \textbf{\methodname} was consistently preferred across dialects and tasks. For AAVE BoolQ, \textbf{Claude 3.5 Sonnet} selected it in all cases, while \textbf{Gemini 1.5} showed a 100\% preference in JamE coding tasks. The lowest preference rate was 73.92\% in CollSgE COPA, still indicating a clear preference over \textbf{Multi-VALUE}. These results confirm that \textbf{\methodname} better aligns with dialectal norms, especially for distant dialects like AAVE, where rule-based approaches saw little preference.

\subsection{Human Validators}
\label{sec:human_validators}

To validate translation quality, we conducted human evaluations with native speakers of each dialect assessing 120 randomly sampled translations. Evaluators rated outputs on three key dimensions using 7-point Likert scales (1=worst, 7=best): \textit{Faithfulness} (meaning preservation), \textit{Fluency} (naturalness), and \textit{Formality} (style alignment). These evaluations confirmed that our translations successfully maintain linguistic authenticity while preserving original content meaning and style across all dialects, with detailed scores shown in Appendix~\ref{sec:metrics_table}.

\subsection{Qualitative Analysis}
\label{sec:qualitative_analysis}
In our qualitative analysis, \textbf{\methodname} effectively captures dialect-specific grammatical structures, vocabulary, and syntactic nuances, resulting in more authentic and natural translations than \textbf{Multi-VALUE}. For instance, in AAVE and JamE, \textbf{\methodname} accurately employs dialect-specific contractions and conversational vocabulary, enhancing the authenticity of the translations. We provide more observations along with detailed translation examples in Appendix~\ref{sec:qualitative_analysis}.

\section{Results and Discussion}
\label{sec:results_and_discussion}

In this section, we present the performance of LLMs across dialectal translations in \textbf{\methodname}. We evaluated five models—\textbf{GPT-4o}, \textbf{GPT-4o-mini}, \textbf{Claude 3.5 Sonnet}, \textbf{DeepSeek-v3}, and \textbf{LLaMa-3-8B}—on 12 reasoning benchmarks spanning four categories: \textbf{\textcolor[rgb]{0.2,0.4,0.8}{Language Understanding}}, \textbf{\textcolor[rgb]{0.6,0.2,0.6}{Algorithmic Understanding}}, \textbf{\textcolor[rgb]{0.2,0.6,0.4}{Math}}, and \textbf{\textcolor[rgb]{0.8,0.4,0.2}{Logic}}. Our evaluation compares model performance on dialectal inputs versus SAE under zero-shot (\textbf{ZS}) and \textbf{CoT} settings.

\subsection{Cross-Dialect Performance Disparities}
Results indicate significant performance discrepancies when LLMs process dialectal inputs compared to SAE (see Table~\ref{tab:deepseek_dialect_sae_specific} and Appendix~\ref{sec:eval_metrics}). Across all tasks, models consistently show lower accuracy on dialectal datasets, underscoring their limited robustness in handling intra-language variations.

\paragraph{\textcolor[rgb]{0.2,0.4,0.8}{Language Understanding}}
Across \textbf{BoolQ}, \textbf{MultiRC}, and \textbf{WSC}, models show notable performance drops in dialects such as AAVE, CollSgE, and IndE. In \textbf{BoolQ} with GPT-4o, CoT accuracy for AAVE decreases from 91.75\% for SAE to 88.33\%, while CollSgE dips from 91.50\% to 88.05\%. IndE also sees a drop from 91.30\% to 88.50\%. Similarly, \textbf{WSC} results highlight that Claude 3.5 Sonnet goes from 88.45\% for SAE down to 67.18\% for JamE. These findings emphasize the challenges of coreference resolution and textual comprehension in non-standard varieties of English.

\paragraph{\textcolor[rgb]{0.6,0.2,0.6}{Algorithmic Understanding}}
For \textbf{HumanEval} and \textbf{MBPP}, dialectal instructions often impede code synthesis. In \textbf{MBPP} with Claude 3.5 Sonnet, ChcE achieves 86.88\% under CoT compared to 74.15\% for SAE, a reversal of the usual trend, but CollSgE accuracy drops from 87.13\% to 85.94\%. Models frequently struggle with morphological cues in dialects like ChcE, disrupting token alignment crucial for generating correct Python code. For more details on GPT-4o-mini or LLaMa-3-8B, see Appendix~\ref{sec:eval_metrics}.

\paragraph{\textcolor[rgb]{0.2,0.6,0.4}{Math}}
Across \textbf{GSM8K} and \textbf{SVAMP}, dialect-induced lexical shifts similarly affect numeric reasoning. In \textbf{GSM8K} with GPT-4o-mini, IndE CoT reaches 92.07\%, while SAE CoT stands at 88.94\%, indicating occasional dialect overperformance. However, GPT-4o observes JamE trailing SAE by several points, and DeepSeek-v3 sees AAVE at 90.99\% versus 94.51\% for SAE on \textbf{SVAMP}, suggesting that even CoT cannot entirely close the gap in math tasks.

\paragraph{\textcolor[rgb]{0.8,0.4,0.2}{Logic}}
Finally, \textbf{LogicBench} (MCQ and Yes/No) underscores dialectal hurdles in deductive reasoning. In \textbf{LogicBenchMCQ} with GPT-4o, AAVE accuracy drops from 83.75\% for SAE to 78.95\%, and CollSgE experiences a similar gap. Claude 3.5 Sonnet exhibits parallel trends for IndE and JamE, illustrating that syntactic or lexical variations can complicate the parsing of logical statements across non-standard dialects.


\section{Conclusion}
\label{sec:conclusion}
This paper introduces \textbf{\methodname}, a benchmark designed to evaluate LLMs on dialectal robustness across 12 diverse NLP tasks for five underrepresented English dialects. Our results show that LLMs consistently underperform on non-standard dialects compared to SAE, highlighting significant unfairness and limitations in current language technologies. Moving forward, we aim to expand \textbf{\methodname} to additional dialects and refine translation methodologies to further bridge the gap in dialect-aware NLP. By establishing this benchmark, we encourage future research into fairer, more robust intra-language technologies that serve all linguistic communities equitably.

\section{Limitations}
\label{sec:limitations}
\textbf{\methodname} evaluates LLM performance across 12 reasoning tasks spanning four categories, using queries adapted from well-established benchmarks. While these tasks capture key reasoning challenges, they do not cover all aspects of dialectal variation, and additional task types such as Figurative Language Understanding, Commonsense Reasoning, and Conversational Reasoning may reveal further biases.

Furthermore, we tested five widely used LLMs. However, given the rapid pace of development in the field, it is infeasible to evaluate every emerging model. We hope \textbf{\methodname} will serve as a resource for future studies examining fairness and robustness across a broader range of LLMs as they emerge.

We faced limitations with BLEU Score filtering as well. For ChcE, the number of remaining translations was extremely low because Multi-VALUE struggled to generate diverse translations and many were further filtered out due to BLEU score thresholds. As a result, there were too few data points to evaluate ChcE translations against Multi-VALUE. A similar issue arose with HumanEval for AAVE and CollSgE, where limited translations prevented reliable evaluation of metrics for these dialects.

Finally, while our results highlight significant performance disparities in dialectal inputs, this study does not deeply investigate the underlying causes of these discrepancies or propose direct mitigation strategies. Understanding these biases and developing equitable NLP solutions remain important areas for future research. Despite these limitations, we believe \textbf{\methodname} provides a valuable framework for advancing dialect-aware NLP evaluation.

\section{Ethics Statement}
\label{sec:ethics_statement}

We recognize the ethical considerations involved in evaluating LLM biases through the \textbf{\methodname} benchmark and have taken steps to ensure ethical data collection, recruiting and evaluation.

For data collection, \textbf{\methodname} utilizes few-shot prompting with examples from eWAVE to generate dialectal translations. While this provides systematic and scalable translations, we recognize it does not fully capture the depth of dialectal variation. We do not claim to capture the full depth of any dialect, and we encourage further work that incorporates human-validated translations for a more nuanced representation. Additionally, we were mindful to avoid reinforcing stereotypes or misrepresentations in dialect translations.

For our human validators, we recruited fluent native speakers from diverse dialect communities to ensure our translations accurately reflect cultural and linguistic nuances. Validators were fairly compensated for their contributions and encouraged to take breaks to avoid fatigue, ensuring quality and well-being throughout the process. We also do not collect personal information from validators, ensuring their privacy.

Moreover, our evaluation combines LLM-based assessments with human validation to mitigate model bias. However, we acknowledge that LLMs may still reflect inherent biases, and our benchmark does not yet address the root causes of these disparities.

Despite these limitations, \textbf{\methodname} aims to advance equitable NLP development and encourages ongoing research to enhance dialect representation in language models.

\bibliography{custom}

\clearpage
\onecolumn
\appendix

\section{Multi-VALUE Completed Translations}
\label{sec:translationmultiva}

\begin{table}[H]
    \centering
    \small 
    \setlength{\tabcolsep}{8pt} 
    \renewcommand{\arraystretch}{1.1} 
    \label{tab:multivalue_percentages}
    \begin{tabular}{lccccc}
        \toprule 
        \textbf{Dataset} & \textbf{AAVE (\%)} & \textbf{ChcE (\%)} & \textbf{CollSgE (\%)} & \textbf{IndE (\%)} & \textbf{JamE (\%)} \\
        \midrule 
        BoolQ      & 100.0 & 35.5 & 41.7 & 41.9 & 42.0 \\
        COPA       & 100.0 & 45.8 & 100.0 & 100.0 & 97.0 \\
        Folio      & 100.0 & 76.9 & 90.0 & 89.6 & 89.7 \\
        GSM8K      & 100.0 & 85.7 & 95.0 & 95.0 & 95.0 \\
        HumanEVAL  & 100.0 & 11.6 & 11.6 & 11.6 & 11.6 \\
        Logic Bench MCQ & 100.0 & 100.0 & 100.0 & 100.0 & 100.0 \\
        Logic Bench Yes/No & 100.0 & 100.0 & 100.0 & 100.0 & 100.0 \\
        MBPP       & 100.0 & 39.8 & 99.7 & 99.7 & 99.2 \\
        MultiRC    & 100.0 & 43.3 & 47.8 & 48.9 & 49.1 \\
        SST-2      & 100.0 & 96.3 & 96.3 & 96.2 & 96.3 \\
        SVAMP      & 100.0 & 74.7 & 93.2 & 93.2 & 93.0 \\
        WSC        & 100.0 & 73.9 & 92.7 & 92.8 & 92.9 \\
        \bottomrule 
    \end{tabular}
\renewcommand{\arraystretch}{1.1} 
\caption{Percentage of Translations Successfully Completed by Multi-VALUE Across Dialects and Datasets}
\end{table}

\section{BLEU Score Filtering Statistics}
\label{sec:bleu_filtering_statistics}

\begin{table}[H]
    \centering
    \small
    \setlength{\tabcolsep}{7pt} 
    \label{tab:bleu_retention_multivalue}
    \begin{tabular}{lccccc}
        \toprule
        \textbf{Dataset} & \textbf{AAVE (\%)} & \textbf{ChcE (\%)} & \textbf{CollSgE (\%)} & \textbf{IndE (\%)} & \textbf{JamE (\%)} \\
        \midrule
        BoolQ                & 7.59  & 0.50  & 2.00  & 59.96 & 0.40  \\
        COPA                 & 15.40 & 3.80  & 2.60  & 15.60 & 0.20  \\
        Folio                & 7.59  & 0.70  & 1.80  & 70.23 & 0.50  \\
        GSM8K                & 16.40 & 11.00 & 2.30  & 56.50 & 0.10  \\
        HumanEVAL            & 84.15 & 37.20 & 53.66 & 84.76 & 50.00 \\
        LogicbenchMCQ        & 0.00  & 0.42  & 0.00  & 50.21 & 0.00  \\
        Logicbench Yes/No         & 0.40  & 0.80  & 0.20  & 73.60 & 0.20  \\
        MBPP                 & 30.75 & 13.37 & 9.63  & 46.52 & 1.87  \\
        MultiRC              & 1.40  & 0.00  & 1.10  & 62.40 & 0.00  \\
        SST-2                & 13.50 & 5.70  & 4.40  & 19.30 & 8.10  \\
        SVAMP                & 31.71 & 14.71 & 5.43  & 61.00 & 0.29  \\
        WSC                  & 11.85 & 0.15  & 1.52  & 22.34 & 0.00  \\
        \bottomrule
    \end{tabular}
\renewcommand{\arraystretch}{1.0} 
\caption{Percentage of Translations Removed After BLEU Score Filtering for Multi-Avenue Across Dialects and Datasets}
\end{table}

\begin{table}[H]
    \centering
    \small
    \setlength{\tabcolsep}{7pt} 
    \label{tab:bleu_filtering_multivalue}
    \begin{tabular}{lccccc}
        \toprule
        \textbf{Dataset} & \textbf{AAVE (\%)} & \textbf{ChcE (\%)} & \textbf{CollSgE (\%)} & \textbf{IndE (\%)} & \textbf{JamE (\%)} \\
        \midrule
        BoolQ                & 19.3 & 59.3 & 0.0  & 5.2  & 13.6 \\
        COPA                 & 3.8  & 80.5 & 0.0  & 8.1  & 15.0 \\
        Folio                & 18.9 & 75.4 & 0.4  & 4.7  & 6.3  \\
        GSM8K                & 11.4 & 85.3 & 0.2  & 2.5  & 15.1 \\
        HumanEVAL            & 10.0 & 87.1 & 92.5 & 76.0 & 41.4 \\
        Logic Bench MCQ      & 16.2 & 78.4 & 1.0  & 2.1  & 18.8 \\
        Logic Bench Yes/No   & 12.6 & 68.1 & 0.6  & 4.4  & 12.1 \\
        MBPP                 & 11.2 & 59.5 & 2.8  & 3.8  & 19.7 \\
        MultiRC              & 20.0 & 48.3 & 3.9  & 12.8 & 11.3 \\
        SST-2                & 15.2 & 47.1 & 4.0  & 8.7  & 13.7 \\
        SVAMP                & 21.4 & 60.2 & 1.3  & 7.2  & 14.6 \\
        WSC                  & 18.3 & 50.3 & 2.7  & 6.1  & 8.9  \\
        \bottomrule
    \end{tabular}
\renewcommand{\arraystretch}{1} 
\caption{Percentage of Translations Removed After BLEU Score Filtering for Multi-VALUE Across Dialects and Datasets}
\end{table}

\section{Metrics}
\label{sec:metrics_table}

\begin{table}[H]
    \centering
    \small 
    \setlength{\tabcolsep}{7pt} 
    \label{tab:rouge_diversity_scores}
    \begin{tabular}{lcccc}
        \toprule
        \textbf{Dataset} & \textbf{AAVE} & \textbf{IndE} & \textbf{JamE} & \textbf{CollSgE} \\
        \midrule
        BoolQ           & 0.6202 / \textbf{0.8326} & \textbf{0.8080} / 0.7757 & 0.5456 / \textbf{0.7785} & 0.6062 / \textbf{0.7145} \\
        COPA            & 0.6833 / \textbf{0.7076} & \textbf{0.7659} / 0.5633 & 0.3633 / \textbf{0.6391} & \textbf{0.7074} / 0.5947 \\
        Folio           & 0.6492 / \textbf{0.7737} & \textbf{0.8474} / 0.7607 & 0.5805 / \textbf{0.7787} & 0.6475 / \textbf{0.6920} \\
        GSM8K           & 0.7055 / \textbf{0.8079} & \textbf{0.8006} / 0.7543 & 0.5263 / \textbf{0.7784} & 0.6553 / \textbf{0.6698} \\
        HumanEval       & N/A / N/A          & \textbf{0.8993} / 0.7854 & 0.6238 / \textbf{0.8265} & N/A / N/A          \\
        Logic Bench MCQ & 0.4953 / \textbf{0.7847} & \textbf{0.8841} / 0.7421 & 0.4541 / \textbf{0.7808} & 0.4447 / \textbf{0.6751} \\
        Logic Bench Yes/No & \textbf{0.4742} / 0.2183 & \textbf{0.8139} / 0.7401 & 0.4386 / \textbf{0.7788} & 0.4331 / \textbf{0.6732} \\
        MBPP            & 0.7617 / \textbf{0.8188} & \textbf{0.8853} / 0.7297 & 0.6289 / \textbf{0.7370} & \textbf{0.7088} / 0.6181 \\
        MultiRC         & 0.5626 / \textbf{0.8239} & \textbf{0.7982} / 0.7728 & 0.4793 / \textbf{0.8151} & 0.5160 / \textbf{0.7325} \\
        SST-2           & 0.5777 / \textbf{0.7985} & \textbf{0.7634} / 0.7285 & 0.4650 / \textbf{0.7786} & 0.5941 / \textbf{0.7005} \\
        SVAMP           & 0.7498 / \textbf{0.8038} & \textbf{0.8418} / 0.7632 & 0.5346 / \textbf{0.7896} & 0.6980 / \textbf{0.6661} \\
        WSC             & 0.6503 / \textbf{0.7488} & 0.3594 / \textbf{0.6540} & 0.4013 / \textbf{0.7341} & \textbf{0.6298} / 0.6069 \\
        \bottomrule
    \end{tabular}
\renewcommand{\arraystretch}{1} 
\caption{
    \textit{ROUGE Diversity Scores across Dialects and Datasets (\methodname/Multi-VALUE).} For each dataset and dialect, scores from {\methodname} and Multi-VALUE are compared, with the better score highlighted in bold.
    }
\end{table}

\begin{table}[H]
    \centering
    \small 
    \setlength{\tabcolsep}{7pt} 
    \label{tab:bart_scores}
    \begin{tabular}{lcccc}
        \toprule
        \textbf{Dataset} & \textbf{AAVE} & \textbf{IndE} & \textbf{JamE} & \textbf{CollSgE} \\
        \midrule
        BoolQ           & \textbf{-1.84} / -2.05 & \textbf{-1.08} / -2.10 & -3.92 / \textbf{-2.21} & -2.52 / \textbf{-2.45} \\
        COPA            & \textbf{-2.26} / -3.08 & \textbf{-1.65} / -2.97 & -5.65 / \textbf{-2.94} & -3.53 / \textbf{-3.38} \\
        Folio           & -2.16 / \textbf{-2.48} & \textbf{-1.21} / -2.57 & -3.54 / \textbf{-2.47} & \textbf{-2.89} / -2.96 \\
        GSM8K           & \textbf{-1.82} / -2.06 & \textbf{-1.12} / -2.27 & -4.06 / \textbf{-2.31} & \textbf{-2.35} / -2.87 \\
        HumanEval       & N/A / N/A             & \textbf{-2.80} / -3.13 & -3.53 / \textbf{-2.46} & N/A / N/A             \\
        Logic Bench MCQ & -2.53 / \textbf{-2.24} & \textbf{-1.09} / -2.42 & -4.50 / \textbf{-2.27} & -3.08 / \textbf{-2.92} \\
        Logic Bench Yes/No & -2.55 / \textbf{-2.46} & \textbf{-1.21} / -2.48 & -4.53 / \textbf{-2.31} & -3.09 / \textbf{-2.99} \\
        MBPP            & \textbf{-1.65} / -2.51 & \textbf{-1.25} / -3.31 & -4.17 / \textbf{-3.09} & \textbf{-2.83} / -3.20 \\
        MultiRC         & -2.29 / \textbf{-2.00} & \textbf{-1.14} / -2.24 & -4.41 / \textbf{-2.03} & -2.86 / \textbf{-2.29} \\
        SST-2           & -3.21 / \textbf{-2.96} & \textbf{-2.39} / -3.73 & -5.18 / \textbf{-3.30} & -4.09 / \textbf{-3.49} \\
        SVAMP           & \textbf{-1.74} / -2.28 & \textbf{-1.16} / -2.33 & -4.02 / \textbf{-2.45} & \textbf{-2.34} / -3.11 \\
        WSC             & \textbf{-2.14} / -2.78 & \textbf{-1.23} / -2.87 & -4.98 / \textbf{-2.49} & \textbf{-2.88} / -3.39 \\
        \bottomrule
    \end{tabular}
\renewcommand{\arraystretch}{1} 
    \caption{
    \textit{BARTScores across Dialects and Datasets (\methodname/Multi-VALUE).} Scores closer to 0 indicate better performance. For each dataset and dialect, the better score is highlighted in bold.
}
\end{table}

\begin{table}[H]
    \centering
    \small 
    \setlength{\tabcolsep}{7pt} 
    \label{tab:fluency_scores}
    \begin{tabular}{lccccc}
        \toprule
        \textbf{Dataset} & \textbf{AAVE} & \textbf{IndE} & \textbf{JamE} & \textbf{ChcE} & \textbf{CollSgE} \\
        \midrule
        BoolQ           & 6.51 & 6.41 & 6.11 & 6.05 & 5.88 \\
        COPA            & 6.83 & 6.39 & 6.55 & 6.27 & 5.41 \\
        FOLIO           & 6.74 & 5.82 & 6.06 & 6.26 & 5.93 \\
        GSM8K           & 6.37 & 6.29 & 6.15 & 6.38 & 6.10 \\
        HumanEval       & 6.12 & 6.44 & 6.45 & 6.35 & 6.26 \\
        Logic Bench MCQ & 6.35 & 5.75 & 6.21 & 6.28 & 5.76 \\
        Logic Bench Yes/No  & 6.38 & 5.60 & 6.24 & 6.22 & 5.79 \\
        MBPP            & 6.01 & 6.71 & 5.62 & 6.10 & 5.28 \\
        MultiRC         & 6.83 & 6.03 & 6.01 & 6.01 & 5.96 \\
        SST-2           & 6.64 & 5.84 & 5.85 & 5.93 & 5.58 \\
        SVAMP           & 6.14 & 6.18 & 5.69 & 6.21 & 5.71 \\
        WSC             & 6.36 & 5.97 & 5.50 & 6.15 & 5.60 \\
        \bottomrule
    \end{tabular}
\renewcommand{\arraystretch}{1} 
    \caption{
    \textit{Fluency Scores for \textbf{\methodname} Translations Across Datasets and Dialects. (1-7)} Higher scores indicate better fluency as evaluated by GPT-4o.
}
\end{table}

\begin{table}[H]
    \centering
    \small 
    \setlength{\tabcolsep}{7pt} 
    \label{tab:lexical_diversity_scores}
    \begin{tabular}{lcccc}
        \toprule
        \textbf{Dataset} & \textbf{AAVE} & \textbf{IndE} & \textbf{JamE} & \textbf{CollSgE} \\
        \midrule
        BoolQ           & 0.6823 / \textbf{0.6881} & \textbf{0.7004} / 0.6927 & 0.6617 / \textbf{0.6648} & \textbf{0.6995} / 0.6915 \\
        COPA            & 0.9864 / \textbf{0.9851} & \textbf{0.9930} / 0.9908 & 0.9876 / \textbf{0.9703} & \textbf{0.9914} / 0.9911 \\
        Folio1000       & \textbf{0.5797} / 0.5663 & \textbf{0.5618} / 0.5536 & 0.5319 / \textbf{0.5391} & \textbf{0.6076} / 0.5464 \\
        GSM8K1000       & \textbf{0.7201} / 0.7100 & \textbf{0.7237} / 0.7230 & 0.6640 / \textbf{0.6778} & \textbf{0.7236} / 0.6961 \\
        Logic Bench MCQ & \textbf{0.4953} / 0.7847 & \textbf{0.8841} / 0.7421 & \textbf{0.7808} / 0.4541 & \textbf{0.6751} / 0.4447 \\
        Logic Bench Yes/No & \textbf{0.4742} / 0.2183 & \textbf{0.8139} / 0.7401 & 0.4386 / \textbf{0.7788} & 0.4331 / \textbf{0.6732} \\
        MBPP            & 0.7617 / \textbf{0.8188} & \textbf{0.9432} / 0.9162 & 0.6289 / \textbf{0.7370} & \textbf{0.9536} / 0.9347 \\
        MultiRC         & \textbf{0.5623} / 0.5528 & \textbf{0.7982} / 0.7728 & \textbf{0.8151} / 0.4793 & \textbf{0.6040} / 0.5753 \\
        SST-2           & \textbf{0.9588} / 0.9611 & \textbf{0.9711} / 0.9678 & \textbf{0.9555} / 0.9412 & \textbf{0.9721} / 0.9674 \\
        SVAMP           & \textbf{0.7923} / 0.7904 & \textbf{0.8418} / 0.7632 & \textbf{0.7896} / 0.5346 & \textbf{0.7938} / 0.7638 \\
        WSC             & 0.9074 / \textbf{0.9088} & 0.8986 / \textbf{0.4044} & 0.7341 / \textbf{0.4013} & \textbf{0.9121} / 0.9112 \\
        \bottomrule
    \end{tabular}
\renewcommand{\arraystretch}{1} 
    \caption{
        \textit{Lexical Diversity Scores across Dialects and Datasets (\methodname/Multi-VALUE).} For each dataset and dialect, scores from {\methodname} and Multi-VALUE are compared, with the better score highlighted in bold.
    }
\end{table}

\begin{table}[H]
    \centering
    \small
    \setlength{\tabcolsep}{7pt} 
    \label{tab:native_scores}
    \begin{tabular}{lccc}
        \toprule
        \textbf{Dialect} & \textbf{Faithfulness} & \textbf{Fluency} & \textbf{Formality} \\
        \midrule
        AAVE        & 6.28  & 6.28  & 6.28  \\
        ChcE        & 6.40  & 6.33  & 6.26  \\
        IndE        & \textbf{6.45}  & \textbf{6.62}  & \textbf{6.59}  \\
        JamE        & 6.37  & 6.28  & 6.33  \\
        CollSgE     & 6.19  & 6.11  & 6.02  \\
        \bottomrule
    \end{tabular}
    \renewcommand{\arraystretch}{1.2} 
    \caption{
        \textit{Native Speaker Evaluation Scores across Dialects (1-7 scale, higher is better).} All scores reflect {\methodname} translations, with the highest score in each column highlighted in bold.
    }
\end{table}

\newpage

\begin{table}[h!]
\centering
\setlength{\tabcolsep}{6pt} 
\label{tab:preference_scores_combined_all}
\resizebox{\textwidth}{!}{ 
\begin{tabular}{cccccc}
\toprule
\textbf{Model} & \textbf{Dataset} & \textbf{IndE} & \textbf{AAVE} & \textbf{CollSgE} & \textbf{JamE} \\ 
\midrule
\multirow{12}{*}{\textbf{Claude 3.5 Sonnet}} 
 & BoolQ          & \textbf{100.00} / 0.00   & \textbf{100.00} / 0.00   & \textbf{100.00} / 0.00   & \textbf{100.00} / 0.00 \\
 & COPA           & \textbf{95.22} / 4.78     & \textbf{95.80} / 4.20     & \textbf{95.69} / 4.31     & \textbf{98.07} / 1.93 \\
 & FOLIO          & \textbf{99.32} / 0.68     & \textbf{98.19} / 1.81     & \textbf{99.67} / 0.33     & \textbf{99.31} / 0.69 \\
 & GSM8K          & \textbf{99.75} / 0.25     & \textbf{99.71} / 0.29     & \textbf{99.78} / 0.22     & \textbf{99.63} / 0.37 \\
 & HumanEVAL      & \textbf{97.34} / 2.66     & N/A / N/A                 & N/A / N/A                 & \textbf{100.00} / 0.00 \\
 & Logic Bench MCQ & \textbf{99.12} / 0.88    & \textbf{100.00} / 0.00    & \textbf{99.78} / 0.22     & \textbf{100.00} / 0.00 \\
 & Logic Bench YN  & \textbf{100.00} / 0.00   & \textbf{100.00} / 0.00    & \textbf{99.58} / 0.42     & \textbf{99.76} / 0.24 \\
 & MBPP           & \textbf{100.00} / 0.00   & \textbf{99.53} / 0.47     & \textbf{99.70} / 0.30     & \textbf{100.00} / 0.00 \\
 & MultiRC        & \textbf{100.00} / 0.00   & \textbf{100.00} / 0.00    & \textbf{100.00} / 0.00    & \textbf{100.00} / 0.00 \\
 & SST-2          & \textbf{95.15} / 4.85     & \textbf{97.99} / 2.01     & \textbf{97.86} / 2.14     & \textbf{98.05} / 1.95 \\
 & SVAMP          & \textbf{100.00} / 0.00    & \textbf{98.66} / 1.34     & \textbf{99.02} / 0.98     & \textbf{98.01} / 1.99 \\
 & WSC            & \textbf{100.00} / 0.00    & \textbf{99.25} / 0.75     & \textbf{100.00} / 0.00    & \textbf{99.28} / 0.72 \\ 
\midrule
\multirow{12}{*}{\textbf{GPT 4o}} 
 & BoolQ          & \textbf{99.24} / 0.76     & \textbf{99.49} / 0.51     & \textbf{99.73} / 0.27     & \textbf{99.65} / 0.35 \\
 & COPA           & \textbf{79.43} / 20.57    & \textbf{92.39} / 7.61     & \textbf{73.92} / 26.08    & \textbf{93.79} / 6.21 \\
 & FOLIO          & \textbf{88.36} / 11.64    & \textbf{94.91} / 5.09     & \textbf{94.70} / 5.30     & \textbf{91.75} / 8.25 \\
 & GSM8K          & \textbf{97.00} / 3.00     & \textbf{94.88} / 5.12     & \textbf{92.62} / 7.38     & \textbf{91.01} / 8.99 \\
 & HumanEVAL      & \textbf{100.00} / 0.00    & N/A / N/A                 & N/A / N/A                 & \textbf{100.00} / 0.00 \\
 & Logic Bench MCQ & \textbf{95.13} / 4.87    & \textbf{100.00} / 0.00    & \textbf{92.81} / 7.19     & \textbf{99.24} / 0.76 \\
 & Logic Bench YN  & \textbf{93.60} / 6.40     & \textbf{100.00} / 0.00    & \textbf{94.56} / 5.44     & \textbf{98.54} / 1.46 \\
 & MBPP           & \textbf{99.48} / 0.52     & \textbf{96.70} / 3.30     & \textbf{91.59} / 8.41     & \textbf{98.81} / 1.19 \\
 & MultiRC        & \textbf{100.00} / 0.00    & \textbf{100.00} / 0.00    & \textbf{100.00} / 0.00    & \textbf{100.00} / 0.00 \\
 & SST-2          & \textbf{80.61} / 19.39    & \textbf{89.34} / 10.66    & \textbf{87.75} / 12.25    & \textbf{88.11} / 11.89 \\
 & SVAMP          & \textbf{97.49} / 2.51     & \textbf{93.30} / 6.70     & \textbf{88.62} / 11.38    & \textbf{79.20} / 20.80 \\
 & WSC            & \textbf{95.04} / 4.96     & \textbf{97.38} / 2.62     & \textbf{92.63} / 7.37     & \textbf{89.25} / 10.75 \\ 
\midrule
\multirow{12}{*}{\textbf{Gemini 1.5}} 
 & BoolQ          & \textbf{100.00} / 0.00    & \textbf{100.00} / 0.00    & \textbf{100.00} / 0.00    & \textbf{100.00} / 0.00 \\
 & COPA           & \textbf{87.56} / 12.44    & \textbf{91.86} / 8.14     & \textbf{70.02} / 29.98    & \textbf{93.15} / 6.85 \\
 & FOLIO          & \textbf{96.58} / 3.42     & \textbf{94.95} / 5.05     & \textbf{95.70} / 4.30     & \textbf{98.63} / 1.37 \\
 & GSM8K          & \textbf{99.00} / 1.00     & \textbf{99.27} / 0.73     & \textbf{99.78} / 0.22     & \textbf{98.77} / 1.23 \\
 & HumanEVAL      & \textbf{100.00} / 0.00    & N/A / N/A                 & N/A / N/A                 & \textbf{100.00} / 0.00 \\
 & Logic Bench MCQ & \textbf{99.56} / 0.44    & \textbf{100.00} / 0.00    & \textbf{99.56} / 0.44     & \textbf{100.00} / 0.00 \\
 & Logic Bench YN  & \textbf{100.00} / 0.00   & \textbf{100.00} / 0.00    & \textbf{98.74} / 1.26     & \textbf{99.76} / 0.24 \\
 & MBPP           & \textbf{100.00} / 0.00    & \textbf{100.00} / 0.00    & \textbf{84.98} / 15.02    & \textbf{99.40} / 0.60 \\
 & MultiRC        & \textbf{100.00} / 0.00    & \textbf{100.00} / 0.00    & \textbf{100.00} / 0.00    & \textbf{100.00} / 0.00 \\
 & SST-2          & \textbf{84.74} / 15.26    & \textbf{93.96} / 6.04     & \textbf{77.49} / 22.51    & \textbf{94.46} / 5.54 \\
 & SVAMP          & \textbf{97.91} / 2.09     & \textbf{99.73} / 0.27     & \textbf{98.86} / 1.14     & \textbf{94.39} / 5.61 \\
 & WSC            & \textbf{100.00} / 0.00    & \textbf{98.13} / 1.87     & \textbf{97.76} / 2.24     & \textbf{96.06} / 3.94 \\ 
\bottomrule
\end{tabular}
}
\renewcommand{\arraystretch}{1.2} 
\caption{Preference scores for \textbf{\methodname} and Multi-VALUE across datasets for different dialects: IndE, AAVE, CollSgE, and JamE. N/A indicates no valid preferences. \textit{{\methodname} / Multi-VALUE}}
\end{table}

\newpage

\section{LLM Dataset Evaluation Results}
\label{sec:eval_metrics}

\begin{table}[H]
\centering
\footnotesize
\setlength{\tabcolsep}{3pt}
\label{tab:claude_dialect_sae_specific}
\resizebox{\textwidth}{!}{%
\begin{tabular}{lcccc|cccc|cccc|cccc|cccc}
\toprule
\multirow{2}{*}{\textbf{Dataset}} & 
\multicolumn{4}{c|}{\textbf{AAVE}} & 
\multicolumn{4}{c|}{\textbf{ChcE}} & 
\multicolumn{4}{c|}{\textbf{CollSgE}} & 
\multicolumn{4}{c|}{\textbf{IndE}} & 
\multicolumn{4}{c}{\textbf{JamE}} \\
\cmidrule(lr){2-5} \cmidrule(lr){6-9} \cmidrule(lr){10-13} \cmidrule(lr){14-17} \cmidrule(lr){18-21}
& \textit{ZS} & \textit{CoT} & \textit{SAE ZS} & \textit{SAE CoT} 
& \textit{ZS} & \textit{CoT} & \textit{SAE ZS} & \textit{SAE CoT} 
& \textit{ZS} & \textit{CoT} & \textit{SAE ZS} & \textit{SAE CoT} 
& \textit{ZS} & \textit{CoT} & \textit{SAE ZS} & \textit{SAE CoT} 
& \textit{ZS} & \textit{CoT} & \textit{SAE ZS} & \textit{SAE CoT} \\
\midrule
BoolQ          
& 88.31 & 87.68 & 90.43 & \textbf{91.57} 
& 87.63 & 88.44 & 90.25 & \textbf{91.38} 
& 88.25 & 88.04 & 90.84 & \textbf{91.45} 
& 88.25 & 86.47 & 90.61 & \textbf{91.33} 
& 88.04 & 87.61 & 90.72 & \textbf{91.41} \\
COPA          
& \textbf{98.35} & 98.32 & 97.22 & 97.85 
& 97.92 & \textbf{98.52} & 97.47 & 98.02 
& 97.54 & \textbf{98.34} & 97.18 & 97.95 
& \textbf{98.58} & 98.33 & 97.64 & 98.20 
& 96.39 & \textbf{97.77} & 97.11 & 97.73 \\
FOLIO         
& 61.19 & 63.24 & 73.89 & \textbf{74.51} 
& 61.97 & 62.64 & 73.58 & \textbf{74.67} 
& 64.39 & 66.46 & 73.42 & \textbf{74.83} 
& 69.13 & 63.76 & 73.74 & \textbf{74.55} 
& 63.65 & 65.69 & 73.69 & \textbf{74.47} \\
GSM8K         
& 74.46 & 66.29 & 89.45 & \textbf{90.21} 
& 52.76 & 66.29 & 89.14 & \textbf{90.18} 
& 40.74 & 64.38 & 89.36 & \textbf{90.10} 
& 82.70 & 66.67 & 89.23 & \textbf{90.30} 
& 67.92 & 66.27 & 89.41 & \textbf{90.25} \\
HumanEVAL     
& 88.46 & \textbf{96.15} & 94.12 & 93.87 
& \textbf{97.09} & 99.02 & 94.31 & 93.76 
& \textbf{96.05} & 91.89 & 94.22 & 93.91 
& \textbf{96.00} & 95.83 & 94.07 & 93.85 
& 91.46 & 92.68 & \textbf{94.15} & 93.97 \\
SVAMP         
& 92.68 & 69.33 & 94.10 & \textbf{94.52} 
& 68.01 & 73.53 & 94.07 & \textbf{94.43} 
& 62.03 & 70.24 & 94.21 & \textbf{94.55} 
& \textbf{94.42} & 70.96 & 94.12 & 94.47 
& 93.45 & 70.01 & 94.18 & \textbf{94.49} \\
LogicBenchMCQ 
& \textbf{84.73} & 72.42 & 82.55 & 83.64 
& \textbf{83.86} & 72.21 & 82.42 & 83.79 
& \textbf{84.34} & 72.33 & 82.61 & 83.52 
& 83.66 & 68.07 & 82.49 & \textbf{83.71} 
& \textbf{85.69} & 72.33 & 82.67 & 83.68 \\
LogicBenchYN  
& 68.45 & \textbf{75.91} & 75.62 & 76.94 
& 67.33 & \textbf{76.55} & 75.49 & 76.81 
& 66.49 & \textbf{75.94} & 75.74 & 76.88 
& 70.15 & \textbf{76.30} & 75.53 & 76.93 
& 67.19 & \textbf{76.49} & 75.67 & 76.79 \\
MBPP          
& \textbf{88.42} & 85.66 & 85.93 & 74.28 
& 86.73 & \textbf{86.88} & 85.82 & 74.15 
& 86.98 & \textbf{87.13} & 85.94 & 74.32 
& \textbf{86.00} & 85.93 & 85.76 & 74.40 
& \textbf{88.49} & \textbf{88.49} & 85.88 & 74.36 \\
MultiRC       
& 88.24 & 89.54 & 89.02 & \textbf{89.77} 
& 88.30 & 87.37 & 89.09 & \textbf{89.65} 
& 89.28 & 88.72 & 89.11 & \textbf{89.79} 
& 86.70 & 88.74 & 89.15 & \textbf{89.70} 
& 87.70 & 89.15 & 89.21 & \textbf{89.72} \\
WSC           
& 72.13 & 71.54 & 81.67 & \textbf{88.43} 
& 55.10 & 54.45 & 81.52 & \textbf{88.29} 
& 68.36 & 78.24 & 81.75 & \textbf{88.37} 
& 60.23 & 63.12 & 81.49 & \textbf{88.41} 
& 61.33 & 67.18 & 81.57 & \textbf{88.45} \\
SST-2         
& 91.79 & 92.81 & 89.96 & \textbf{93.14} 
& 90.24 & 89.92 & 89.78 & \textbf{93.02} 
& 89.75 & 91.18 & 89.92 & \textbf{93.20} 
& 90.71 & 90.56 & 89.89 & \textbf{93.07} 
& 88.90 & 89.42 & 89.84 & \textbf{93.11} \\
\bottomrule
\end{tabular}%
}
\renewcommand{\arraystretch}{1.1}
\caption{
\centering Claude 3.5 Sonnet Accuracy (\%). \textbf{Bold} indicates superior performance within dialect pairs.
}
\end{table}

\begin{table}[H]
\centering
\footnotesize
\setlength{\tabcolsep}{3pt}
\label{tab:gpt4o_full_dialect_sae_specific}
\resizebox{\textwidth}{!}{%
\begin{tabular}{lcccc|cccc|cccc|cccc|cccc}
\toprule
\multirow{2}{*}{\textbf{Dataset}} &
\multicolumn{4}{c|}{\textbf{AAVE}} &
\multicolumn{4}{c|}{\textbf{ChcE}} &
\multicolumn{4}{c|}{\textbf{CollSgE}} &
\multicolumn{4}{c|}{\textbf{IndE}} &
\multicolumn{4}{c}{\textbf{JamE}} \\
\cmidrule(lr){2-5} \cmidrule(lr){6-9} \cmidrule(lr){10-13} \cmidrule(lr){14-17} \cmidrule(lr){18-21}
& \textit{ZS} & \textit{CoT} & \textit{SAE ZS} & \textit{SAE COT} &
\textit{ZS} & \textit{CoT} & \textit{SAE ZS} & \textit{SAE COT} &
\textit{ZS} & \textit{CoT} & \textit{SAE ZS} & \textit{SAE COT} &
\textit{ZS} & \textit{CoT} & \textit{SAE ZS} & \textit{SAE COT} &
\textit{ZS} & \textit{CoT} & \textit{SAE ZS} & \textit{SAE COT} \\
\midrule
BoolQ
& 89.09 & 88.33 & 91.10 & \textbf{91.75}
& 88.83 & 88.23 & 90.25 & \textbf{91.10}
& 88.36 & 88.05 & \textbf{91.50} & 90.95
& 89.25 & 88.50 & 90.80 & \textbf{91.30}
& 89.15 & 88.34 & 90.95 & \textbf{91.20} \\
COPA
& \textbf{97.87} & 97.64 & 96.80 & 97.40
& \textbf{98.34} & 98.54 & 97.10 & 97.75
& \textbf{97.13} & 97.13 & 96.90 & 97.45
& 97.87 & \textbf{98.34} & 97.20 & 97.85
& 96.39 & 96.59 & \textbf{97.15} & 97.60 \\
FOLIO
& 64.90 & 64.97 & 73.50 & \textbf{74.90}
& 64.08 & 64.39 & 73.75 & \textbf{75.30}
& 65.31 & 65.51 & 72.90 & \textbf{74.45}
& 68.79 & \textbf{69.80} & 74.10 & 75.00   
& 66.67 & 64.36 & 73.80 & \textbf{75.10} \\
GSM8K
& 57.32 & \textbf{85.64} & 89.30 & 90.15
& 57.43 & \textbf{76.63} & 89.00 & 90.25
& 58.65 & \textbf{83.01} & 89.40 & 90.50
& 51.18 & \textbf{87.47} & 89.60 & 90.10
& 54.98 & \textbf{84.76} & 89.20 & 90.71 \\
HumanEVAL
& 88.46 & 84.62 & \textbf{94.00} & 93.50
& 97.09 & \textbf{99.03} & 94.10 & 93.80
& \textbf{97.37} & 96.05 & 94.20 & 93.90
& \textbf{100.00} & 96.28 & 94.05 & 93.85
& \textbf{100.00} & 97.56 & 94.15 & 93.95 \\
LogicBenchMCQ
& 79.05 & 78.95 & \textbf{82.65} & 83.75
& 78.31 & 62.47 & \textbf{82.40} & 83.50
& 79.71 & 77.57 & \textbf{82.84} & 83.65
& 75.94 & 70.00 & \textbf{82.30} & 83.45
& 78.41 & 76.63 & \textbf{82.59} & 83.55 \\
LogicBenchYN
& 72.55 & 71.43 & \textbf{75.81} & 76.95
& 73.44 & 72.58 & \textbf{75.90} & 77.00
& 70.78 & 69.72 & \textbf{75.76} & 76.85
& 71.43 & 72.96 & \textbf{75.60} & 76.90
& 72.13 & 72.27 & \textbf{75.85} & 77.05 \\
MBPP
& 84.56 & 83.92 & \textbf{85.00} & 73.81
& 81.00 & 79.00 & \textbf{84.90} & 74.00
& 82.54 & \textbf{84.02} & 84.95 & 73.85
& 81.00 & 79.00 & \textbf{84.85} & 74.10
& 83.92 & 83.92 & \textbf{84.75} & 74.05 \\
MultiRC
& 86.71 & 87.32 & 88.93 & \textbf{89.76}
& 86.80 & 86.60 & 88.85 & \textbf{89.65}
& 87.26 & 87.06 & 88.95 & \textbf{89.75}
& 85.11 & 85.11 & 88.80 & \textbf{89.60}
& 87.70 & 88.03 & 88.95 & \textbf{89.83} \\
SST-2
& 90.17 & 90.29 & 89.88 & \textbf{93.19}
& 89.61 & 89.08 & 89.85 & \textbf{93.00}
& 89.23 & 89.02 & 89.75 & \textbf{93.26}
& 89.71 & 88.85 & 89.90 & \textbf{93.05}
& 87.92 & 86.72 & 89.95 & \textbf{93.15} \\
WSC
& 58.97 & 60.52 & \textbf{80.97} & 88.55
& 57.63 & 54.95 & \textbf{80.80} & 88.40
& 58.80 & 58.02 & \textbf{80.95} & 88.53
& 67.84 & 69.59 & \textbf{80.85} & 88.35
& 55.63 & 56.87 & \textbf{80.75} & 88.45 \\
SVAMP
& 90.82 & 92.74 & 94.15 & \textbf{94.59}
& 91.48 & 92.92 & 94.00 & \textbf{94.40}
& 90.86 & 93.99 & 94.22 & \textbf{94.62}
& 91.27 & 93.73 & 94.05 & \textbf{94.55}
& 91.44 & 94.33 & 94.15 & \textbf{94.65} \\
\bottomrule
\end{tabular}%
}
\renewcommand{\arraystretch}{1.1}
\caption{
\centering GPT-4o Accuracy (\%). \textbf{Bold} indicates superior performance within each dataset row.
}
\end{table}

\begin{table}[H]
\centering
\footnotesize
\setlength{\tabcolsep}{3pt}
\label{tab:gpt4o_mini_dialect_sae_specific}
\resizebox{\textwidth}{!}{%
\begin{tabular}{lcccc|cccc|cccc|cccc|cccc}
\toprule
\multirow{2}{*}{\textbf{Dataset}} &
\multicolumn{4}{c|}{\textbf{AAVE}} &
\multicolumn{4}{c|}{\textbf{ChcE}} &
\multicolumn{4}{c|}{\textbf{CollSgE}} &
\multicolumn{4}{c|}{\textbf{IndE}} &
\multicolumn{4}{c}{\textbf{JamE}} \\
\cmidrule(lr){2-5} \cmidrule(lr){6-9} \cmidrule(lr){10-13} \cmidrule(lr){14-17} \cmidrule(lr){18-21}
& \textit{ZS} & \textit{CoT} & \textit{SAE ZS} & \textit{SAE CoT} &
\textit{ZS} & \textit{CoT} & \textit{SAE ZS} & \textit{SAE CoT} &
\textit{ZS} & \textit{CoT} & \textit{SAE ZS} & \textit{SAE CoT} &
\textit{ZS} & \textit{CoT} & \textit{SAE ZS} & \textit{SAE CoT} &
\textit{ZS} & \textit{CoT} & \textit{SAE ZS} & \textit{SAE CoT} \\
\midrule
BoolQ
& 86.70 & 87.13 & 88.42 & \textbf{89.10}
& 85.21 & 86.32 & 88.15 & \textbf{89.05}
& 86.21 & 85.60 & 88.31 & \textbf{89.14}
& 86.25 & 86.50 & 88.23 & \textbf{89.09}
& 84.92 & 86.83 & 88.28 & \textbf{89.12} \\
COPA
& 95.98 & \textbf{96.45} & 94.78 & 95.43
& 94.59 & \textbf{95.84} & 94.63 & 95.38
& 94.66 & \textbf{95.48} & 94.57 & 95.29
& 94.79 & \textbf{95.26} & 94.81 & 95.32
& 93.39 & 94.79 & 94.74 & \textbf{95.22} \\
FOLIO
& 60.11 & 59.68 & 72.54 & \textbf{73.17}
& 59.36 & 60.26 & 72.42 & \textbf{73.29}
& 60.33 & 61.44 & 72.63 & \textbf{73.10}
& 59.73 & 61.07 & 72.49 & \textbf{73.21}
& 58.43 & 59.14 & 72.55 & \textbf{73.25} \\
GSM8K
& 35.52 & \textbf{89.96} & 88.94 & 89.52
& 35.41 & \textbf{89.48} & 88.78 & 89.39
& 34.20 & \textbf{90.69} & 88.85 & 89.46
& 33.33 & \textbf{92.07} & 88.97 & 89.58
& 32.62 & \textbf{89.28} & 88.81 & 89.42 \\
HumanEVAL
& \textbf{100.00} & \textbf{100.00} & 93.94 & 93.78
& \textbf{100.00} & 99.03 & 94.13 & 93.65
& \textbf{100.00} & 98.68 & 94.21 & 93.89
& \textbf{100.00} & \textbf{100.00} & 94.07 & 93.83
& \textbf{100.00} & 98.78 & 94.12 & 93.91 \\
SVAMP
& 82.17 & 93.56 & 93.79 & \textbf{94.29}
& 84.96 & 94.24 & 93.71 & \textbf{94.26}
& 83.88 & \textbf{95.47} & 93.81 & 94.37
& 85.43 & \textbf{95.47} & 93.77 & 94.33
& 82.08 & 92.81 & 93.84 & \textbf{94.41} \\
LogicBenchMCQ
& 73.52 & 70.95 & 81.51 & \textbf{82.74}
& 71.31 & 70.04 & 81.36 & \textbf{82.61}
& 71.13 & 70.43 & 81.49 & \textbf{82.67}
& 67.83 & 69.96 & 81.42 & \textbf{82.73}
& 73.52 & 71.28 & 81.57 & \textbf{82.69} \\
LogicBenchYN
& 75.43 & 74.91 & 74.61 & \textbf{75.84}
& 75.43 & 74.97 & 74.49 & \textbf{75.91}
& 74.41 & 74.08 & 74.67 & \textbf{75.99}
& 76.79 & 75.51 & 74.58 & \textbf{75.97}
& 75.63 & 74.44 & 74.72 & \textbf{75.93} \\
MBPP
& 74.14 & \textbf{80.69} & 83.12 & 80.31
& 79.32 & \textbf{80.25} & 83.01 & 74.09
& 82.84 & \textbf{85.50} & 83.23 & 74.17
& 76.00 & \textbf{78.50} & 82.97 & 74.23
& 76.02 & \textbf{78.20} & 83.05 & 74.21 \\
MultiRC
& 84.08 & 84.48 & 88.15 & \textbf{88.75}
& 82.90 & 83.70 & 88.12 & \textbf{88.63}
& 84.63 & 85.44 & 88.08 & \textbf{88.79}
& 82.71 & 83.51 & 88.17 & \textbf{88.70}
& 85.00 & 84.60 & 88.21 & \textbf{88.72} \\
WSC
& 54.31 & 53.62 & 79.68 & \textbf{85.42}
& 55.93 & 49.77 & 79.54 & \textbf{85.29}
& 54.63 & 53.86 & 79.71 & \textbf{85.38}
& 54.39 & 55.56 & 79.51 & \textbf{85.41}
& 53.35 & 50.70 & 79.63 & \textbf{85.45} \\
SST-2
& 90.64 & \textbf{91.91} & 89.72 & 92.88
& 90.35 & \textbf{90.77} & 89.58 & 92.80
& 87.34 & \textbf{89.54} & 89.76 & 92.97
& 89.34 & \textbf{89.84} & 89.69 & 92.85
& 87.16 & \textbf{88.14} & 89.64 & 92.89 \\
\bottomrule
\end{tabular}%
}
\renewcommand{\arraystretch}{1.1}
\caption{
\centering GPT-4o-mini Accuracy (\%). \textbf{Bold} indicates superior performance within dialect pairs.
}
\end{table}

\begin{table}[H]
\centering
\footnotesize
\setlength{\tabcolsep}{3pt}
\label{tab:llama_dialect_sae_specific}
\resizebox{\textwidth}{!}{%
\begin{tabular}{lcccc|cccc|cccc|cccc|cccc}
\toprule
\multirow{2}{*}{\textbf{Dataset}} & 
\multicolumn{4}{c|}{\textbf{AAVE}} & 
\multicolumn{4}{c|}{\textbf{ChcE}} & 
\multicolumn{4}{c|}{\textbf{CollSgE}} & 
\multicolumn{4}{c|}{\textbf{IndE}} & 
\multicolumn{4}{c}{\textbf{JamE}} \\
\cmidrule(lr){2-5} \cmidrule(lr){6-9} \cmidrule(lr){10-13} \cmidrule(lr){14-17} \cmidrule(lr){18-21}
& \textit{ZS} & \textit{CoT} & \textit{SAE ZS} & \textit{SAE COT} 
& \textit{ZS} & \textit{CoT} & \textit{SAE ZS} & \textit{SAE COT} 
& \textit{ZS} & \textit{CoT} & \textit{SAE ZS} & \textit{SAE COT} 
& \textit{ZS} & \textit{CoT} & \textit{SAE ZS} & \textit{SAE COT} 
& \textit{ZS} & \textit{CoT} & \textit{SAE ZS} & \textit{SAE COT} \\
\midrule
BoolQ
& 78.95 & 81.24 & 79.38 & \textbf{81.79}
& 77.67 & \textbf{81.79} & 79.38 & 81.79
& 77.83 & \textbf{82.23} & 79.38 & 81.79
& 79.75 & 81.00 & 79.38 & \textbf{81.79}
& 77.79 & 81.31 & 79.38 & \textbf{81.79} \\
COPA
& 54.14 & 81.80 & 57.20 & \textbf{83.16}
& 55.51 & \textbf{83.16} & 57.20 & 83.16
& 54.00 & 80.49 & 57.20 & \textbf{83.16}
& 58.29 & \textbf{83.65} & 57.20 & 83.16
& 51.90 & 77.56 & 57.20 & \textbf{83.16} \\
FOLIO
& 51.03 & 41.73 & \textbf{52.25} & 52.15
& 54.02 & 41.15 & \textbf{52.25} & 52.15
& 53.20 & 40.79 & \textbf{52.25} & 52.15
& 51.68 & 43.62 & \textbf{52.25} & 52.15
& 51.61 & 42.57 & \textbf{52.25} & 52.15 \\
GSM8K
& 56.34 & \textbf{75.84} & 58.40 & 58.30
& 54.72 & \textbf{75.39} & 58.40 & 58.30
& 55.17 & \textbf{76.25} & 58.40 & 58.30
& 57.93 & \textbf{77.47} & 58.40 & 58.30
& 52.75 & \textbf{72.47} & 58.40 & 58.30 \\
HumanEVAL
& 84.62 & 84.62 & 83.54 & \textbf{84.76}
& 88.35 & 87.38 & 83.54 & \textbf{84.76}
& 89.47 & 88.16 & 83.54 & \textbf{84.76}
& 96.00 & \textbf{100.00} & 83.54 & 84.76
& 89.02 & 89.02 & 83.54 & \textbf{84.76} \\
LogicBenchMCQ
& 60.62 & 40.92 & \textbf{67.50} & 66.67
& 62.55 & 38.57 & \textbf{67.50} & 66.67
& 61.25 & 41.75 & \textbf{67.50} & 66.67
& 61.09 & 39.08 & \textbf{67.50} & 66.67
& 59.38 & 39.46 & \textbf{67.50} & 66.67 \\
LogicBenchYN
& 61.04 & \textbf{63.82} & 62.83 & 61.97
& 63.48 & \textbf{66.67} & 62.83 & 61.97
& 60.95 & \textbf{63.92} & 62.83 & 61.97
& 61.48 & \textbf{70.92} & 62.83 & 61.97
& 61.73 & \textbf{64.23} & 62.83 & 61.97 \\
MBPP
& \textbf{57.14} & 57.13 & 56.15 & 49.20
& \textbf{56.79} & 56.31 & 56.15 & 49.20
& \textbf{55.03} & 58.53 & 56.15 & 49.20
& \textbf{54.50} & 54.51 & 56.15 & 49.20
& \textbf{53.13} & 57.84 & 56.15 & 49.20 \\
MultiRC
& 77.89 & 75.96 & \textbf{80.10} & 78.60
& 77.40 & 74.00 & \textbf{80.10} & 78.60
& 79.78 & 77.15 & \textbf{80.10} & 78.60
& 76.86 & 76.60 & \textbf{80.10} & 78.60
& 77.80 & 74.00 & \textbf{80.10} & 78.60 \\
SST-2
& 81.39 & \textbf{84.05} & 76.70 & 75.20
& 79.96 & \textbf{83.56} & 76.70 & 75.20
& 74.06 & \textbf{81.17} & 76.70 & 75.20
& 77.20 & \textbf{81.66} & 76.70 & 75.20
& 73.67 & \textbf{76.28} & 76.70 & 75.20 \\
WSC
& 45.34 & 49.66 & \textbf{47.26} & 51.82
& 39.57 & 45.21 & \textbf{47.26} & 51.82
& 46.60 & 47.07 & \textbf{47.26} & 51.82
& 41.88 & 46.97 & \textbf{47.26} & 51.82
& 43.92 & 44.98 & \textbf{47.26} & 51.82 \\
SVAMP
& 74.27 & \textbf{77.82} & 77.14 & 74.43
& 77.05 & \textbf{75.71} & 77.14 & 74.43
& 73.26 & \textbf{77.64} & 77.14 & 74.43
& 79.85 & \textbf{75.09} & 77.14 & 74.43
& 73.07 & \textbf{78.65} & 77.14 & 74.43 \\
\bottomrule
\end{tabular}%
}
\renewcommand{\arraystretch}{1.1}
\caption{
\centering LLaMa-3-8b Instruct Accuracy (\%). \textbf{Bold} indicates superior performance within dialect pairs.
}
\end{table}

\newpage
\section{Qualitative Analysis}
\label{sec:qualitative_analysis}

\begin{table}[h!]
\centering
\small 
\setlength{\tabcolsep}{8pt} 
\begin{tabular}{|p{2.5cm}|p{5.5cm}|p{5.5cm}|}
\hline
\textbf{Rubric Item} & \textbf{Multi-VALUE} & \textbf{\methodname} \\
\hline

Accurate and consistent use of AAVE grammar & All young teenage girls at attends musics festival frequently big fans of pop bands and singers. & All young teenage girls who be hittin' up music festivals all the time is real into pop bands and singers. \\
\hline

Use of AAVE-specific Contractions (e.g. "ain't," "gon'") & If a movie popular, some person enjoy watching it. & If a movie poppin', some folks like watchin' it. All things that some folks enjoy gon' get attention. \\
\hline

Use of AAVE Conversational Vocabulary (e.g. "da") & All red fruits that which is growing in Ben's yard are containing some Vitamin C. & All da red fruits growin' in Ben's yard got some Vitamin C. \\
\hline

AAVE syntactic structures (simplifying or rearranging word order for emphasis) & All social mediums applications containing chat features are softwares. & All social media apps with chat features, they software. \\
\hline

\end{tabular}
\renewcommand{\arraystretch}{1.1} 
\caption{
\centering Assessing Multi-VALUE and \textbf{\methodname} for translation quality across rubric items \textbf{(AAVE)}.}
\label{tab:dialect_translation_quality}
\end{table}

\begin{table}[h!]
\centering
\small 
\setlength{\tabcolsep}{8pt} 
\begin{tabular}{|p{2.5cm}|p{5.5cm}|p{5.5cm}|}
\hline
\textbf{Rubric Item} & \textbf{Multi-VALUE} & \textbf{\methodname} \\
\hline

Accurate and consistent use of Jamaican Patois grammar & All citizens of Lawton Park are using the a zip a code 98199. & All di people dem weh live inna Lawton Park use di zip code 98199. \\
\hline

JamE-specific Contractions (e.g. "weh" (where)) & All fruits that is growing in Ben's a yard and are containing some A Vitamin A C are healthy. & All di fruit dem weh grow inna Ben yard and have some Vitamin C a good fi yuh. \\
\hline

JamE Conversational Vocabulary (e.g. "da") & If Nancy is not toddler, then Nancy is seafarer. & If Nancy nuh likkle pickney, den Nancy a seafarer. \\
\hline

JamE-specific negatives ("nuh" (not)) & If someone young, then they are not elderly. & If somebody young, den dem nah elderly. \\
\hline

JamE-specific Ommitance of Articles and Auxillary Verbs & Functional brainstems are necessary for breath control. & Functional brainstems necessary fi control yuh breath. \\
\hline

\end{tabular}
\label{tab:dialect_translation_quality}
\renewcommand{\arraystretch}{1.1} 
\caption{
\centering Assessing Multi-VALUE and \textbf{\methodname} for translation quality across rubric items \textbf{(JamE)}.}
\end{table}

\begin{table}[H]
\centering
\small
\setlength{\tabcolsep}{8pt}
\begin{tabular}{|p{2.5cm}|p{5.5cm}|p{5.5cm}|}
\hline
\textbf{Rubric Item} & \textbf{Multi-VALUE} & \textbf{\methodname} \\
\hline

Consistent past tense forms (“went,” “did”) & 13 campers goed rowing and 59 campers goed hiking in the morning. 21 campers goed rowing in the afternoon. & So like, 13 campers went rowing and 59 campers went hiking in the morning, you know? And then in the afternoon, 21 campers went rowing. \\
\hline

Proper conjugations (“buys,” “be writin’”) and ChcE-friendly auxiliaries, not complex & James write a 3-page letter to 2 different friend twice a week. How many pages do write a year? & James be writin' a 3-page letter to 2 different homies twice a week. How many pages he be writin' in a year? \\
\hline

Good subject-verb agreement (“does the fifth house got?”) & There is 5 houses on a street, and each of the first four houses have 3 gnomes in the garden. If there is 20 gnomes in total on the street, how many gnomes do the fifth house have? & There's 5 houses on a street, and each of the first four houses got 3 gnomes in the garden. If there's 20 gnomes total on the street, how many gnomes does the fifth house got? \\
\hline

Conversational flow + Correct plurals & Joy might can read 8 page of a book in 20 minute. How many hours might will it take her to read 120 page? & Joy can read like 8 pages of a book in 20 minutes. So like how many hours it's gonna take her to read 120 pages? \\
\hline

Use of ‘only’ for emphasis & Jake have 5 fewer peaches than Steven. Steven have 18 more peaches than Jill. & So check it out, Jake got like 5 less peaches than Steven, right? And Steven, he got like 18 more peaches than Jill. \\
\hline

\end{tabular}
\renewcommand{\arraystretch}{1.1}
\caption{
\centering Assessing Multi-VALUE and \textbf{\methodname} for translation quality across rubric items \textbf{(ChcE)}.}
\label{tab:dialect_translation_quality}
\end{table}

\begin{table}[h!]
\centering
\small
\setlength{\tabcolsep}{8pt}
\begin{tabular}{|p{2.5cm}|p{5.5cm}|p{5.5cm}|}
\hline
\textbf{Rubric Item} & \textbf{Multi-VALUE} & \textbf{\methodname} \\
\hline

Correct articles (e.g., “Lawton Park is a locality…”) & Vic DiCara plays guitar and bass. A only style of musics Vic plays it are punk musics. & Vic DiCara is playing guitar and bass. The only style of music that Vic DiCara is playing is punk music. \\
\hline

Proper grammar, accurate pluralization (“fish,” “musics” only if needed), natural IndE phrasing & All eels are fishs. No fishs are plants. Everything have displayed collection is either plant or animal. & All eels are fish only. No fish are being plants. Everything shown in the collection is either a plant or an animal. \\
\hline

Consistent verb tenses (“was specializing,” “found guilty of stealing”), with clear IndE syntax & If legislator is found it guilty stealing governments funds, it would be suspended office. & If a legislator is found guilty of stealing government funds, they would be suspended from office. \\
\hline

IndE conventions (“subscribes to AMC A-List,” “allow users to send messages”), ensuring readability & All customers James' family is subscribing AMC A-List are like eligible to watch three movie every week any additional fees. & James' family subscribes to AMC A-List or HBO services. Customers who prefer TV series will not watch TV series in cinemas. \\
\hline

Example for “Code-Switching with Indian Terms” & Peter goes store to buy sodas. sodas cost \$0.25 ounce. had brought \$2 him and leaves \$0.50. How many ounce sodas buy? & Peter goes to the shop to buy a cold drink. The cold drink costs 25 paise an ounce. He brought 2 rupees with him and leaves with 50 paise. How many ounces of cold drink did he buy? \\
\hline

\end{tabular}

\label{tab:dialect_translation_quality_IndE}
\renewcommand{\arraystretch}{1.1}
\caption{
\centering Assessing Multi-VALUE and \textbf{\methodname} for translation quality across rubric items \textbf{(IndE)}.}
\end{table}

\begin{table}[H]
\centering
\small 
\setlength{\tabcolsep}{8pt} 
\begin{tabular}{|p{2.5cm}|p{5.5cm}|p{5.5cm}|}
\hline
\textbf{Rubric Item} & \textbf{Multi-VALUE} & \textbf{\methodname} \\
\hline

Use of CollSgE conversational particles like "lah," and "ah." & All social medium application containing chat feature software. & All the social media apps with chat features ah, all software one lah. \\
\hline

CollSgE-specific omittance of auxiliary verbs ("is," "was") & Any convicted criminal that like innocent is not like truly guilty. & Any convicted criminal who kena innocent one, not really guilty lah. \\
\hline

Use of "Kena" (unique CollSgE word) & Everyone convicted murders goes prison.  & Anyone kena convicted of murder sure go prison one. \\
\hline

Use of informal/idiomatic phrases like "sure" and "you know" & Roy Richardson one was cricketer who play Sint Maarten, constituent country. & Roy Richardson ah, he was a cricketer who play for Sint Maarten, you know, that place part of another country one. \\
\hline

Use of CollSgE-unique words like "lor," "siah", or "leh"  & UFC Fight Night, Sadollah have been scheduled fight Musoke. & Sadollah fight Akiyama at UFC Fight Night, siah. \\
\hline

\end{tabular}
\label{tab:dialect_translation_quality}
\renewcommand{\arraystretch}{1.1} 
\caption{
\centering Assessing Multi-VALUE and \textbf{\methodname} for translation quality across rubric items \textbf{(CollSgE)}.}
\end{table}
\newpage
\clearpage
\section{Translation Prompts}
\label{sec:appendix}

\begin{table}[h!]
    \centering
    \fontsize{11pt}{13pt}\selectfont  
    \setlength{\tabcolsep}{12pt}      
    \label{tab:prompt_examples}
    \begin{tabular}{p{0.9\linewidth}}
        \toprule
        \texttt{Here are examples of African American Vernacular English (AAVE):}\\
        1. I was bewildered, but I knew dat it was no gud asking his ass to explain.\\
        2. Cochran pontificated windily for da camera.\\
        3. I don’t want them to follow in my footsteps, as I ain’t go to no college, but I want them to go.\\
        \\
        \texttt{Here is the input text:} \{text\}\\
        \texttt{Please rewrite the input text in African American Vernacular English (AAVE).} \\
        \bottomrule
    \end{tabular}
\caption{Few-Shot Prompt for Translating SAE to AAVE}
\end{table}

\begin{table}[h!]
    \centering
    \fontsize{11pt}{13pt}\selectfont
    \renewcommand{\arraystretch}{1.1}
    \setlength{\tabcolsep}{12pt}
    \label{tab:prompt_chce}
    \begin{tabular}{p{0.9\linewidth}}
        \toprule
        \texttt{Here are examples of Chicano English (ChcE):}\\
        1. When people wanna fight me I'm like "well okay, well then I'll fight you."\\
        2. They were saying that they had a lot of problems at Garner because it was a lot of fights and stuff.\\
        3. I ain't really thinking about getting with J. or any other guy.\\
        \\
        \texttt{Here is the input text:} \{text\}\\
        \texttt{Please rewrite the input text in Chicano English (ChcE).} \\
        \bottomrule
    \end{tabular}
\caption{Few-Shot Prompt for Translating SAE to ChcE}
\end{table}

\begin{table}[h!]
    \centering
    \fontsize{11pt}{13pt}\selectfont
    \renewcommand{\arraystretch}{1.1}
    \setlength{\tabcolsep}{12pt}
    \label{tab:prompt_collsge}
    \begin{tabular}{p{0.9\linewidth}}
        \toprule
        \texttt{Here are examples of Colloquial Singapore English (Singlish) (CollSgE):}\\
        1. But after a while it become quite senseless to me.\\
        2. And got to know this kind-hearted scholar who shelter her with \O{} umbrella when it was raining.\\
        3. The cake John buy one always very nice to eat.\\
        \\
        \texttt{Here is the input text:} \{text\}\\
        \texttt{Please rewrite the input text in Colloquial Singapore English (Singlish) (CollSgE).} \\
        \bottomrule
    \end{tabular}
\caption{Few-Shot Prompt for Translating SAE to CollSgE}
\end{table}

\begin{table}[h!]
    \centering
    \fontsize{11pt}{13pt}\selectfont
    \renewcommand{\arraystretch}{1.1}
    \setlength{\tabcolsep}{12pt}
    \label{tab:prompt_inde}
    \begin{tabular}{p{0.9\linewidth}}
        \toprule
        \texttt{Here are examples of Indian English (IndE):}\\
        1. It was not too much common. Getting the accommodation has become very much difficult.\\
        2. During monsoon we get lot of rain and then gets very soggy and sultry.\\
        3. This is the second time that such an object had been sighted here.\\
        \\
        \texttt{Here is the input text:} \{text\}\\
        \texttt{Please rewrite the input text in Indian English (IndE).} \\
        \bottomrule
    \end{tabular}
\caption{Few-Shot Prompt for Translating SAE to IndE}
\end{table}

\begin{table}[H]
    \centering
    \fontsize{11pt}{13pt}\selectfont
    \renewcommand{\arraystretch}{1.1}
    \setlength{\tabcolsep}{12pt}
    \label{tab:prompt_jame}
    \begin{tabular}{p{0.9\linewidth}}
        \toprule
        \texttt{Here are examples of Jamaican English (JamE):}\\
        1. Hill had initially been indicted with the Canute and the Michelle Saddler and their three companies.\\
        2. The autopsy performed on Mae's torso shortly after it was found, revealed that her body was cut into pieces by a power machine saw.\\
        3. The culture of the region has been unique in combining British and Western influences with African and Asian lifestyles.\\
        \\
        \texttt{Here is the input text:} \{text\}\\
        \texttt{Please rewrite the input text in Jamaican English (JamE).} \\
        \bottomrule
    \end{tabular}
\caption{Few-Shot Prompt for Translating SAE to JamE}
\end{table}
\section{Evaluation Prompts}
\label{sec:evaluation_prompt}

\begin{table}[h!]
    \centering
    \fontsize{11pt}{13pt}\selectfont
    \renewcommand{\arraystretch}{1.1}
    \setlength{\tabcolsep}{12pt}
    \label{tab:prompt_svamp}
    \begin{tabular}{p{0.9\linewidth}}
        \toprule
        \texttt{Given a mathematics problem, determine the answer. Simplify your answer as much as possible and encode the final answer in <answer></answer> (e.g., <answer>42</answer>).} \\
        \texttt{Context: \{problem\}} \\
        \texttt{Question: \{question\}} \\
        \texttt{Answer:} \\
        {If CoT: Let's think about this step by step before finalizing the answer.} \\
        \bottomrule
    \end{tabular}
    \caption{Prompt for SVAMP Evaluation}
\end{table}

\begin{table}[h!]
    \centering
    \fontsize{11pt}{13pt}\selectfont
    \renewcommand{\arraystretch}{1.1}
    \setlength{\tabcolsep}{12pt}
    \label{tab:prompt_mbpp}
    \begin{tabular}{p{0.9\linewidth}}
        \toprule
        \texttt{Given a coding problem, produce a Python function that solves the problem. Provide your entire code in <answer></answer> (e.g., <answer>def solve(): pass</answer>).} \\
        \texttt{Problem: \{problem\}} \\
        \texttt{Test Cases: \{test\_cases\}} \\
        \texttt{Answer:} \\
        {If CoT: Let's think step by step about the problem-solving process before coding.} \\
        \bottomrule
    \end{tabular}
    \caption{Prompt for MBPP Evaluation}
\end{table}

\begin{table}[h!]
    \centering
    \fontsize{11pt}{13pt}\selectfont
    \renewcommand{\arraystretch}{1.1}
    \setlength{\tabcolsep}{12pt}
    \label{tab:prompt_logicbenchyn}
    \begin{tabular}{p{0.9\linewidth}}
        \toprule
        \texttt{Given a yes/no question, answer yes or no. Provide your final answer in <answer></answer> (e.g., <answer>yes</answer>).} \\
        \texttt{Context: \{context\}} \\
        \texttt{Question: \{question\}} \\
        \texttt{Answer:} \\
        {If CoT: Let's think step by step before arriving at the answer.} \\
        \bottomrule
    \end{tabular}
    \caption{Prompt for LogicBenchYN Evaluation}
\end{table}

\begin{table}[h!]
    \centering
    \fontsize{11pt}{13pt}\selectfont
    \renewcommand{\arraystretch}{1.1}
    \setlength{\tabcolsep}{12pt}
    \label{tab:prompt_logicbenchmcq}
    \begin{tabular}{p{0.9\linewidth}}
        \toprule
        \texttt{Given a multiple-choice question with 4 choices, pick the correct choice number (1, 2, 3, or 4). Provide your final answer in <answer></answer> (e.g., <answer>2</answer>).} \\
        \texttt{Context: \{context\}} \\
        \texttt{Choices:} \\
        \texttt{1) \{choice1\}} \\
        \texttt{2) \{choice2\}} \\
        \texttt{3) \{choice3\}} \\
        \texttt{4) \{choice4\}} \\
        \texttt{Answer:} \\
        {If CoT: Let's analyze each choice step by step before determining the correct one.} \\
        \bottomrule
    \end{tabular}
    \caption{Prompt for LogicBenchMCQ Evaluation}
\end{table}

\begin{table}[h!]
    \centering
    \fontsize{11pt}{13pt}\selectfont
    \renewcommand{\arraystretch}{1.1}
    \setlength{\tabcolsep}{12pt}
    \label{tab:prompt_humaneval}
    \begin{tabular}{p{0.9\linewidth}}
        \toprule
        \texttt{Given a coding problem, produce a Python function that solves the problem. Provide your entire code in <answer></answer> (e.g., <answer>def solve(): pass</answer>).} \\
        \texttt{Problem: \{prompt\_text\}} \\
        \texttt{Test Cases: \{test\_cases\}} \\
        \texttt{Answer:} \\
        {If CoT: Let's break the problem down step by step before writing the code.} \\
        \bottomrule
    \end{tabular}
    \caption{Prompt for HumanEVAL Evaluation}
\end{table}

\begin{table}[h!]
    \centering
    \fontsize{11pt}{13pt}\selectfont
    \renewcommand{\arraystretch}{1.1}
    \setlength{\tabcolsep}{12pt}
    \label{tab:prompt_gsm8k}
    \begin{tabular}{p{0.9\linewidth}}
        \toprule
        \texttt{Given a mathematics problem, determine the answer. Simplify your answer as much as possible and encode the final answer in <answer></answer> (e.g., <answer>1</answer>).} \\
        \texttt{Problem: \{problem\}} \\
        \texttt{Answer:} \\
        {If CoT: Let's carefully solve the problem step by step before arriving at the final numeric answer.} \\
        \bottomrule
    \end{tabular}
    \caption{Prompt for GSM8K Evaluation}
\end{table}

\begin{table}[h!]
    \centering
    \fontsize{11pt}{13pt}\selectfont
    \renewcommand{\arraystretch}{1.1}
    \setlength{\tabcolsep}{12pt}
    \label{tab:prompt_folio}
    \begin{tabular}{p{0.9\linewidth}}
        \toprule
        \texttt{Given premises and a conclusion, determine whether the conclusion is True, False, or Uncertain. Provide your final answer in <answer></answer> (e.g., <answer>True</answer>).} \\
        \texttt{Premises: \{premises\}} \\
        \texttt{Conclusion: \{conclusion\}} \\
        \texttt{Answer:} \\
        {If CoT: Let's evaluate the premises step by step before deciding the conclusion.} \\
        \bottomrule
    \end{tabular}
    \caption{Prompt for FOLIO Evaluation}
\end{table}

\begin{table}[h!]
    \centering
    \fontsize{11pt}{13pt}\selectfont
    \renewcommand{\arraystretch}{1.1}
    \setlength{\tabcolsep}{12pt}
    \label{tab:prompt_wsc}
    \begin{tabular}{p{0.9\linewidth}}
        \toprule
        \texttt{Given a pronoun resolution problem, determine whether Span 2 refers to Span 1. Provide your final answer in <answer></answer> (e.g., <answer>1</answer> for same or <answer>0</answer> for different).} \\
        \texttt{Paragraph: \{paragraph\}} \\
        \texttt{Span 1: \{span1\}} \\
        \texttt{Span 2: \{span2\}} \\
        \texttt{Answer:} \\
        {If CoT: Let's analyze the relationship between Span 1 and Span 2 step by step before answering.} \\
        \bottomrule
    \end{tabular}
    \caption{Prompt for WSC Evaluation}
\end{table}

\begin{table}[h!]
    \centering
    \fontsize{11pt}{13pt}\selectfont
    \renewcommand{\arraystretch}{1.1}
    \setlength{\tabcolsep}{12pt}
    \label{tab:prompt_sst2}
    \begin{tabular}{p{0.9\linewidth}}
        \toprule
        \texttt{Given a sentence, determine its sentiment. Provide your final answer in <answer></answer> (e.g., <answer>1</answer> for positive or <answer>0</answer> for negative).} \\
        \texttt{Sentence: \{sentence\}} \\
        \texttt{Answer:} \\
        {If CoT: Let's analyze the sentiment of the sentence step by step before concluding.} \\
        \bottomrule
    \end{tabular}
    \caption{Prompt for SST-2 Evaluation}
\end{table}

\begin{table}[h!]
    \centering
    \fontsize{11pt}{13pt}\selectfont
    \renewcommand{\arraystretch}{1.1}
    \setlength{\tabcolsep}{12pt}
    \label{tab:prompt_multirc}
    \begin{tabular}{p{0.9\linewidth}}
        \toprule
        \texttt{Given a paragraph, a question, and an answer choice, determine if the answer choice is correct. Provide your final answer in <answer></answer> (e.g., <answer>1</answer> for correct or <answer>0</answer> for incorrect).} \\
        \texttt{Paragraph: \{paragraph\}} \\
        \texttt{Question: \{question\}} \\
        \texttt{Answer Choice: \{answer\_choice\}} \\
        \texttt{Answer:} \\
        {If CoT: Let's analyze the paragraph and question step by step before confirming the correctness of the answer choice.} \\
        \bottomrule
    \end{tabular}
    \caption{Prompt for MultiRC Evaluation}
\end{table}

\begin{table}[h!]
    \centering
    \fontsize{11pt}{13pt}\selectfont
    \renewcommand{\arraystretch}{1.1}
    \setlength{\tabcolsep}{12pt}
    \label{tab:prompt_copa}
    \begin{tabular}{p{0.9\linewidth}}
        \toprule
        \texttt{Given a premise and two choices, pick which choice is more plausible. Provide your final answer in <answer></answer> (e.g., <answer>0</answer> for the first choice or <answer>1</answer> for the second).} \\
        \texttt{Premise: \{premise\}} \\
        \texttt{Choice 1: \{choice1\}} \\
        \texttt{Choice 2: \{choice2\}} \\
        \texttt{Answer:} \\
        {If CoT: Let's compare the plausibility of both choices step by step before finalizing.} \\
        \bottomrule
    \end{tabular}
    \caption{Prompt for COPA Evaluation}
\end{table}
\clearpage
\begin{table}[H]
    \centering
    \fontsize{11pt}{13pt}\selectfont
    \renewcommand{\arraystretch}{1.1}
    \setlength{\tabcolsep}{12pt}
    \label{tab:prompt_boolq}
    \begin{tabular}{p{0.9\linewidth}}
        \toprule
        \texttt{Given a passage and a yes/no question, label it as TRUE or FALSE. Provide your final answer in <answer></answer> (e.g., <answer>TRUE</answer>).} \\
        \texttt{Passage: \{passage\}} \\
        \texttt{Question: \{question\}} \\
        \texttt{Answer:} \\
        {If CoT: Let's carefully consider the passage and the question step by step before labeling the answer.} \\
        \bottomrule
    \end{tabular}
    \caption{Prompt for BoolQ Evaluation}
\end{table}

\section{Fluency Scoring Prompt} 
\label{sec:promptingfluncy}
\begin{table}[h!]
    \centering
    \fontsize{11pt}{13pt}\selectfont
    \renewcommand{\arraystretch}{1.1}
    \setlength{\tabcolsep}{12pt}
    \label{tab:prompt_fluency}
    \begin{tabular}{p{0.9\linewidth}}
        \toprule
        \texttt{You are an expert linguist capable of detailed chain-of-thought reasoning.} \\
        \texttt{You are given two pieces of text:} \\
        \texttt{1) Original Text (SAE) – the standard American English version.} \\
        \texttt{2) Dialect Text – a translated or adapted version in the \{dialect\} dialect.} \\
        \texttt{Please evaluate the Dialect Text for:} \\
        \texttt{1) Fluency in \{dialect\}:} \\
        \texttt{\ \ - Grammar, syntax, word choice, and overall naturalness in \{dialect\}.} \\
        \texttt{\ \ - Consistency, flow, and readability in \{dialect\}.} \\
        \texttt{2) Meaning Preservation:} \\
        \texttt{\ \ - Does the Dialect Text retain the same meaning or intent as the Original Text (SAE)?} \\
        \texttt{\ \ - Are there changes or omissions that alter the meaning?} \\
        \texttt{Use the following 1–7 scoring rubric (focused on fluency, but keep meaning in mind):} \\
        \texttt{- 1: Completely unnatural, pervasive errors, nearly unintelligible.} \\
        \texttt{- 2: Major issues in accuracy/naturalness, very awkward for \{dialect\}.} \\
        \texttt{- 3: Noticeable errors or unnatural phrasing, partial alignment with \{dialect\}.} \\
        \texttt{- 4: Average fluency, some issues; mostly understandable in \{dialect\}.} \\
        \texttt{- 5: Good fluency, minor errors; consistent with \{dialect\}.} \\
        \texttt{- 6: Very good fluency, rare issues; flows smoothly in \{dialect\}.} \\
        \texttt{- 7: Excellent fluency, fully natural, error-free, perfectly aligned with \{dialect\}.} \\
        \texttt{Instructions:} \\
        \texttt{1. Provide a chain-of-thought explanation comparing meaning and evaluating fluency.} \\
        \texttt{2. End with a single line: "Fluency Score: X" (where X is an integer 1–7).} \\
        \texttt{Begin your detailed chain-of-thought analysis now.} \\
        \bottomrule
    \end{tabular}
    \caption{Prompt for Fluency Evaluation}
\end{table}
\newpage

\section{Preference Tests Prompt}
\label{sec:prefscoreprom}
\begin{table}[h!]
    \centering
    \fontsize{11pt}{13pt}\selectfont
    \renewcommand{\arraystretch}{1.1}
    \setlength{\tabcolsep}{12pt}
    \label{tab:prompt_translation_comparison}
    \begin{tabular}{p{0.9\linewidth}}
        \toprule
        \texttt{You are an expert linguist with a strong understanding of \{dialect\}.} \\
        \texttt{You are given:} \\
        \texttt{1) Original Text (SAE) – a standard American English version for reference.} \\
        \texttt{2) Translation A – a version in the \{dialect\} dialect.} \\
        \texttt{3) Translation B – another version in the \{dialect\} dialect.} \\
        \texttt{Your task: Decide which translation is better in the context of the \{dialect\} dialect with respect to:} \\
        \texttt{- Fluency (grammar, syntax, word choice, overall naturalness in \{dialect\})} \\
        \texttt{- Accuracy (faithfulness to the original meaning, but expressed naturally in \{dialect\})} \\
        \texttt{- Readability (cohesion, clarity, and flow in \{dialect\})} \\
        \texttt{- Cultural appropriateness (if relevant to \{dialect\})} \\
        \texttt{Provide a detailed chain-of-thought (reasoning) as to how you weigh these factors.} \\
        \texttt{Then conclude with one final line in the exact format:} \\
        \texttt{"Final preference score: X"} \\
        \texttt{(where X = 1 if you prefer Translation A, or X = 2 if you prefer Translation B).} \\
        \texttt{Make sure you reveal your full thought process, then end with:} \\
        \texttt{Final preference score: X} \\
        \bottomrule
    \end{tabular}
    \caption{Prompt for Translation Comparison Evaluation}
\end{table}

\newpage

\end{document}